%% file: root.tex
\documentclass[runningheads]{llncs}

\usepackage[T1]{fontenc}

\usepackage{graphicx}
\usepackage{comment}
\usepackage{amsmath,amssymb}
\usepackage{color}
\usepackage{url}

\usepackage{authblk}
\usepackage{cite}
\usepackage{amsfonts}
\usepackage{algorithmic}
\usepackage{textcomp}
\usepackage[flushleft]{threeparttable}
\usepackage{comment}
\usepackage{multirow}
\usepackage[dvipsnames]{xcolor}
\usepackage{subcaption}
\usepackage{dblfloatfix}
\usepackage{float}
\usepackage{arydshln}
\usepackage[export]{adjustbox}

\usepackage{hyperref}

\newcommand\possg[1]{{\color{ForestGreen}#1}} 
\newcommand\posns[1]{{\color{Gray}#1}} 
\newcommand\negsg[1]{{\color{Maroon}#1}} 
\newcommand\negns[1]{{\color{Gray}#1}} 

\newif\ifreview
\reviewfalse

\ifreview
	\usepackage{lineno}

	\linenumbers
\fi

\begin{document}


\def\SubNumber{041}

\def\GCPRTrack{Main Track}

\title{RC-BEVFusion: A Plug-In Module for Radar-Camera Bird's~Eye~View~Feature~Fusion}

\ifreview
\else
	\titlerunning{RC-BEVFusion: Radar-Camera Bird's~Eye~View~Feature~Fusion}

	\author{Lukas~St\"acker\inst{1,2}\orcidID{0000-0001-7537-606X} \and
	Shashank Mishra\inst{2,3}\orcidID{0000-0003-2882-2789} \and
	Philipp~Heidenreich\inst{1}\orcidID{0000-0002-5906-7125} \and
	Jason~Rambach\inst{3}\orcidID{0000-0001-8122-6789} \and
	Didier~Stricker\inst{2,3}\orcidID{0000-0002-5708-6023}}
	
	\authorrunning{L. St\"acker et al.}
	
	\institute{Stellantis, Opel Automobile GmbH, R\"usselsheim am Main, Germany \and Rheinland-Pf\"alzische Technische Universit\"at Kaiserslautern-Landau, Kaiserslautern, Germany \and German Research Center for Artificial Intelligence, Kaiserslautern, Germany}
\fi

\maketitle              

\begin{abstract}
Radars and cameras belong to the most frequently used sensors for advanced driver assistance systems and automated driving research. However, there has been surprisingly little research on radar-camera fusion with neural networks. One of the reasons is a lack of large-scale automotive datasets with radar and unmasked camera data, with the exception of the nuScenes dataset. Another reason is the difficulty of effectively fusing the sparse radar point cloud on the bird's eye view (BEV) plane with the dense images on the perspective plane. The recent trend of camera-based 3D object detection using BEV features has enabled a new type of fusion, which is better suited for radars. In this work, we present RC-BEVFusion, a modular radar-camera fusion network on the BEV plane. We propose two novel radar encoder branches, and show that they can be incorporated into several state-of-the-art camera-based architectures. We show significant performance gains of up to 28\% increase in the nuScenes detection score, which is an important step in radar-camera fusion research. Without tuning our model for the nuScenes benchmark, we achieve the best result among all published methods in the radar-camera fusion category.
\end{abstract}

\section{Introduction}

\begin{figure*}[bt]
\captionsetup{font=small}
\centering
  \includegraphics[width=\textwidth]{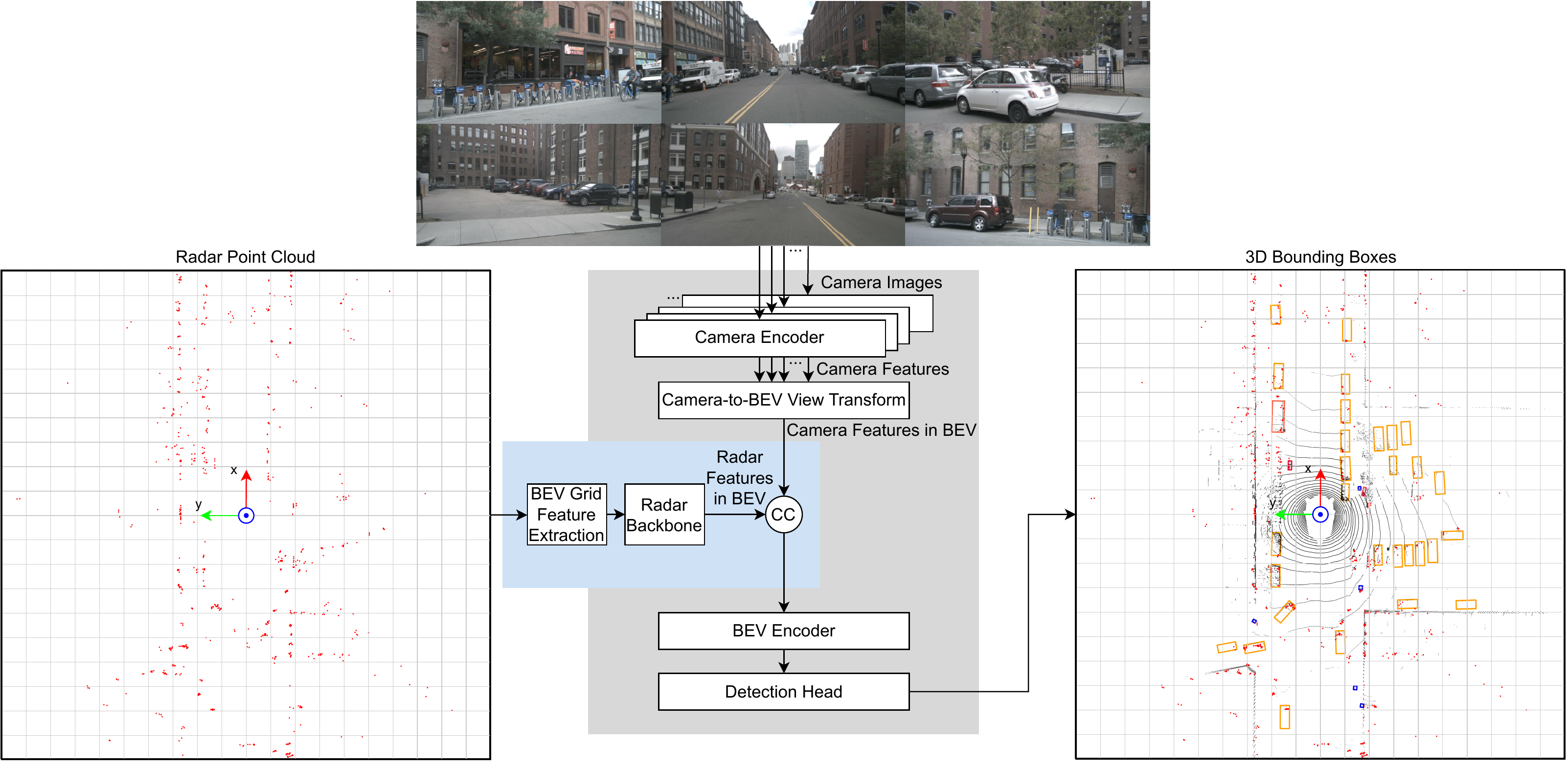}
  \caption{Overview of RC-BEVFusion network architecture. The block marked in grey is inherited from an exchangeable camera-only baseline, while the block marked in blue shows our proposed radar-camera fusion plug-in module.}
  \label{fig:network_architecture}
\end{figure*}

The development of advanced driver assistance systems (ADAS) and automated driving functions has made remarkable progress in recent years, resulting in increased safety and convenience for drivers. A robust environment perception is the key requirement for these systems, which rely on sensors such as radar, camera, or LiDAR to detect surrounding objects. Each sensor has its own advantages and disadvantages that must be considered when designing a perception system. An extensive review on multi-modular automotive object detection is presented in \cite{DiFeng.2020}. 

Radar sensors are advantageous in that they are less affected by adverse environmental conditions such as rain, fog, or darkness, and they have a longer detection range when compared to cameras and LiDARs. However, they are limited in their ability to provide detailed information about the shape and texture of objects \cite{Zhou.2022}. Cameras, on the other hand, provide rich visual information and can recognize objects based on their appearance, but their performance can be affected by changes in lighting conditions and inaccurate depth estimation \cite{Ma.2021, Hung.2022}. LiDARs provide detailed 3D information and are less affected by lighting conditions, but they can be expensive and have limited range \cite{Zhou.2022}.

Sensor fusion has the potential to overcome these limitations of individual sensors. In particular, the combination of radar and camera sensors arguably offers the most complementary features. The main challenge is how to associate radar and camera features given that conventional radars provide data on the bird’s eye view (BEV) plane, whereas cameras provide data on the image plane. Projecting radar points to the image discards too much geometric information, whereas projecting camera features to the sparse radar points discards too much semantic information \cite{Liu.2022}. 

Recent advancements in camera-only networks using view transformers \cite{Philion.2020, Huang.2021} have enabled a new type of fusion on the BEV plane, which is well suited for radar data. In this paper, we propose RC-BEVFusion, a novel radar-camera fusion architecture on the BEV plane inspired by \cite{Liu.2022} and illustrated in Figure~\ref{fig:network_architecture}. In contrast to previous radar-camera fusion techniques \cite{Nabati.2021, Kim.2022, Pang.2023}, our architecture allows radar and camera features to equally contribute to the final detections, enabling the network to detect obstacles that may be missed by one of the modalities. It is a flexible architecture that inherits several elements from an exchangable camera-only baseline: a camera encoder, a camera-to-BEV view transformer, a BEV encoder and a detection head. On top of these modules, we propose two radar encoder branches: RadarGridMap and BEVFeatureNet. Our results show that they can be used as a plug-in module in various camera-based architectures and significantly enhance their performance.

To train and evaluate our network, we need a large-scale automotive dataset with radar point clouds, unmasked camera images with a lot of variety in the scenes and 3D object annotations. A recent overview on radar datasets is given in \cite{Zhou.2022}. First, there are datasets with conventional 2+1D radar sensors that provide a list of detections with measurements of the range, range rate, azimuth angle, and radar cross section (RCS). Out of these, the nuScenes dataset \cite{Caesar.2019} is the only one that fulfils our requirements. Second, there are recent datasets with high-performance 3+1D radar sensors that provide denser point clouds and additionally measure the elevation angle. From this group, the Astyx \cite{Meyer.2019} dataset is too small, while the  View-of-Delft \cite{Palffy.2022} dataset is medium-sized but has limited visual variety in the images due to the high annotation frequency. The recently presented TJ4DRadSet \cite{Zheng.2022} may be a future option but the data is not fully released yet. In this work, we therefore choose to conduct our experiments with the nuScenes dataset \cite{Caesar.2019}. 

The remaining paper is organized as follows. Section~\ref{section:2} provides an overview of related work on automotive object detection with camera-only, radar-only and radar-camera fusion. Section~\ref{section:3} describes the proposed radar-camera fusion architecture on the BEV plane. Section~\ref{section:4} presents extensive experimental results on the nuScenes dataset and demonstrates the effectiveness of the proposed architecture. Finally, Section~\ref{section:5} concludes the paper and outlines future work.

\section{Related work}
\label{section:2}

The task of 3D object detection is mostly conducted with the help of cameras, LiDARs and, less frequently, radars. In the following, we give an overview of recent advances on image-only and radar-only object detection, as well as LiDAR-camera and radar-camera sensor fusion.

\subsection{Image-only object detection}
Image-based 3D object detection is a difficult task in the field of computer vision because it involves identifying and localizing objects in 3D space using only a single camera as a sensor. This is in contrast to LiDAR and radar systems, which provide depth measurements and can more accurately determine the 3D location of objects.

Early approaches to image-based 3D object detection focused on using known geometric information to estimate 3D bounding boxes from 2D detections \cite{Mousavian.2017}. More recent techniques have extended existing 2D object detection models with additional detection heads specifically designed for 3D object detection \cite{Simonelli.2019, Wang.2021, Zhou.2019}. Some approaches have also used predicted depth maps as auxiliary features \cite{Manhardt.2019} or to create a pseudo-LiDAR point cloud, which is then processed using LiDAR-based object detectors \cite{Wang.2019c}.

The most recent research in this area has mainly followed two directions. The first is the use of transformer-based techniques, which leverage the ability of transformer models to process sequences of data and perform self-attention to learn more complex relationships between features \cite{Wang.2022, Liu.2022b}. The second direction is the development of BEV-based object detectors. To this end, the features need to be transformed from the image plane to the BEV plane. A pioneering work uses orthographic feature transform \cite{Roddick.2018}, where a voxel grid is projected to the image to extract features. To reduce memory consumption, the voxel grid is then collapsed along the vertical axis to create BEV features. More recent methods are based on the Lift-Splat-Shoot (LSS) view transformer \cite{Philion.2020}, which lifts image features into a 3D pseudo point cloud via dense depth prediction, before again collapsing the vertical dimension to create BEV features. The idea is first incorporated into a 3D object detection network in \cite{Huang.2021}, and later refined with LiDAR-based depth supervision \cite{Li.2022b} and temporal stereo \cite{Li.2022c}. In \cite{Zhou.2022c}, a more efficient view transformer in terms of memory and computations is introduced by formulating LSS into matrix operations with decomposed ring and ray matrices, and compressing image features and the estimated depth map along the vertical dimension. In this work, we build upon these BEV-based object detectors and integrate our radar-camera fusion as a plug-in module.

\subsection{Radar-only object detection}
Automotive radars are a common sensor used in autonomous vehicles for detecting objects in the environment. These radars typically provide a preprocessed list of detections, but the sparsity and lack of semantic information make it difficult to use the data for stand-alone 3D object detection. As a result, much of the research in this area has focused on either semantic segmentation of radar point clouds \cite{Schumann.2018} or experimental setups using the raw radar cube \cite{Major.2019, Palffy.2020}.

Recently, there have been some point cloud-based techniques for object detection as well. These can be broadly divided into convolutional and graph-based approaches. Some convolutional approaches assign each point in the point cloud to a cell in a BEV grid and include feature layers such as the maximum RCS and Doppler values \cite{Dreher.2020, Scheiner.2021}. Since the BEV grid is similar in structure to an image, traditional convolutional networks can be applied for object detection. Other convolutional approaches use variants of PointPillars \cite{Lang.2019} to create a pillar grid automatically from the point cloud \cite{Scheiner.2021, Palffy.2022}. In contrast, graph neural network based approaches perform object detection directly on the radar point cloud \cite{Danzer.2019, Scheiner.2020}. Recent work combines both approaches by first extracting features with a graph neural network and then mapping the features to a BEV grid for further processing \cite{Ulrich.2022}. In this work, we examine radar feature encoders inspired by these ideas.

\subsection{Sensor fusion object detection}

Sensor fusion aims at leveraging the strengths of a diverse sensor combination. Most sensor fusion research focuses on LiDAR-camera fusion, as LiDAR provides accurate 3D information and cameras provide high semantic value, while sharing the same optical propagation principles.
There are several techniques for fusing LiDAR and camera data. Some approaches are based on projecting 2D detections from the camera into a frustum and matching them with LiDAR points to refine the 3D detections \cite{Qi.2018, Wang.2019b}. Other techniques augment the LiDAR points with semantics from the image and use LiDAR-based object detectors to perform detection \cite{Vora.2020, Wang.2021b, Fei.2020}.
The most recent approaches to LiDAR-camera fusion involve extracting BEV features from both the LiDAR and camera data and fusing them on the BEV plane before applying a joint BEV encoder to perform object detection \cite{Liu.2022, Liang.2022, Drews.2022}.

When compared to LiDAR, automotive radar uses a different wavelength and measurement principle. As a consequence, radar typically shows strong RCS fluctuations and reduced resolution in range and angle, resulting in less dense point clouds. Moreover, whereas modern radars also measure elevation, many conventional radars only provide detections on the BEV plane. These differences make the fusion especially challenging and prevent the simple replacement of LiDAR with radar processing.
Early research commonly projected the radar detections onto the image plane to associate the data. This approach can be used to find regions of interest in the image \cite{Nabati.2019, Nabati.2020} or to create additional image channels with radar data that can be used with image-based networks \cite{Nobis.2019, Chadwick.2019, Stacker.2022}.

More recent methods have moved away from this 2D approach and instead focus on fusing based on 3D information. One approach is to refine image-based 3D detections with associated radar data \cite{Nabati.2021}. This can be sub-optimal because it discards the possibility of radar-only detections. Another approach is to project 3D regions of interest (ROIs) to the image and BEV plane to extract features from each sensor \cite{Kim.2020}. Finally, cross-attention has been used to align and fuse the features in 3D \cite{Hwang.2022, Pang.2023}. In this work, we propose a novel architecture to fuse radar and camera data on the BEV plane.

\section{RC-BEVFusion}
\label{section:3}

In this section, we describe our proposed model architectures. We start by giving an overview of the general fusion architecture, before providing more detail on the proposed radar encoders, the camera-only networks we use as baselines and the loss function.

\subsection{Overview}

In this work, we introduce a novel radar branch and use it as a plug-in module on different camera-based 3D object detection networks to improve their performance. The prerequisite for our proposed radar-camera fusion is that the camera-only network uses BEV features as an intermediate representation. The general architecture is shown in Figure~\ref{fig:network_architecture}. The block marked in grey is inherited from an exchangeable camera-only baseline, while the block marked in blue shows our proposed radar plug-in module.

First, BEV features are extracted from the images and the radar point cloud separately. To this end, a backbone network is used to extract features from each image before they are transformed into joint BEV features with a view transformer. We set up our radar encoder so that it creates BEV features in the same shape and geometric orientation as the camera BEV features. The features are then fused by concatenation followed by a 1$\times$1 convolutional layer that reduces the embedding dimension to the original dimension of the camera BEV features. We can then use the same BEV encoder and 3D detection heads that are used in the respective camera-only network, introducing little overhead with our fusion.

\subsection{Radar encoders}

\begin{figure*}[bt]
\captionsetup{font=small}
\centering
  \includegraphics[width=0.8\textwidth]{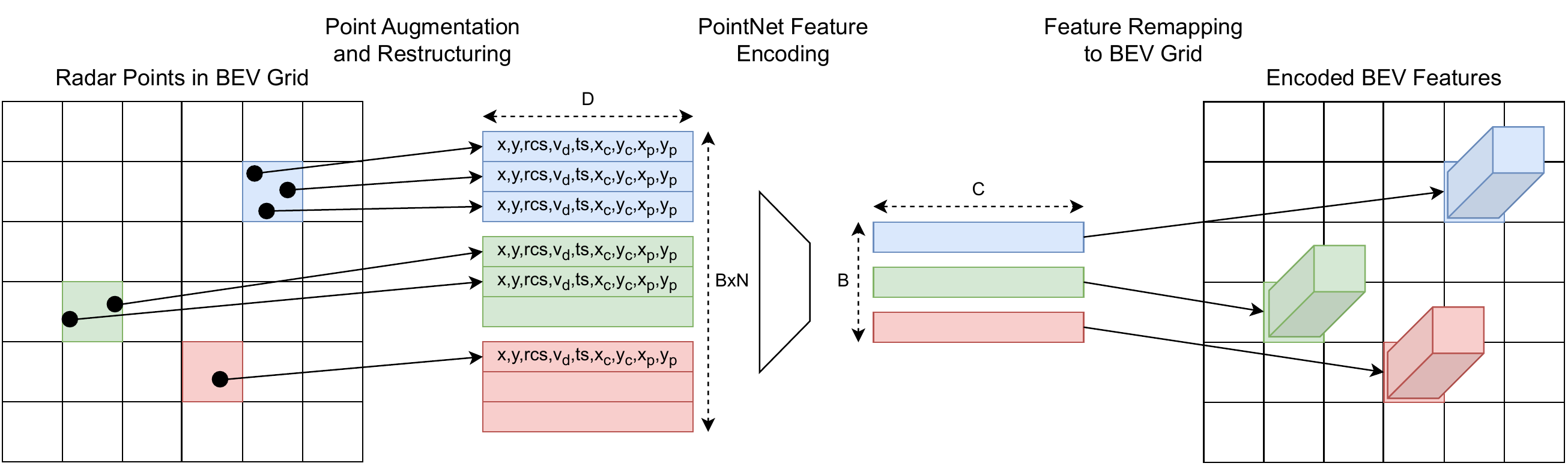}
  \caption{BEVFeatureNet radar encoder. The radar detections are mapped to BEV grid cells for augmentation and restructuring into a dense tensor. After applying a simplified PointNet, the encoded features are remapped to the BEV grid.
  }
  \label{fig:bevfeaturenet}
\end{figure*}

We propose two radar encoders, RadarGridMap and BEVFeatureNet, which we explain in the following. They each consist of two stages: First, we create a regular, structured BEV grid from the sparse radar point cloud. Then, we apply a convolutional backbone that further encodes the BEV features.

\subsubsection{RadarGridMap}

Inspired by \cite{Dreher.2020}, we design a hand-crafted radar BEV grid. We map each detection to a cell on the grid and fill the cell with four channels: the number of detections per cell, the maximum RCS value, and the minimum and maximum signed compensated Doppler values. 

After the grid mapping, we use a small generalized ResNet \cite{He.2016} as our radar backbone. We use 16 layers grouped into residual blocks with BatchNorm and ReLU. We use two downsampling stages that double the channels but reduce the resolution of the BEV grid. We design the size of the BEV grid so that the output of the radar backbone has the same shape as the camera BEV features.

\subsubsection{BEVFeatureNet}
The BEVFeatureNet illustrated in Figure~\ref{fig:bevfeaturenet} is inspired by the pillar feature encoding of PointPillars \cite{Lang.2019}, but adapted for radar data. First, we map each point of the radar point cloud to a cell in a predefined BEV grid. In the original pillar feature encoding, each point in the LiDAR point cloud has coordinates $x$, $y$, and $z$, and reflectivity $r$. Further, the point is augmented with the distances to the arithmetic mean of all points in its pillar, $x_c$, $y_c$, and $z_c$ and the offsets to the pillar center $x_p$ and $y_p$. The 2+1D radar point cloud in nuScenes \cite{Caesar.2019} does not have a $z$-coordinate or a reflectivity $r$, but instead it has a radial velocity $v_d$, measured via the Doppler effect, and an $\mathrm{RCS}$. We therefore discard the $z$-axis and its augmented feature, replace the reflectivity with the $\mathrm{RCS}$, and include the radial velocity values. We further aggregate multiple radar sweeps and append the timestamp difference to the latest sweep $t_s$ to each point. Thus, we obtain the 9-dimensional feature set:
\begin{equation}
    \Vec{F} = [x, y, \mathrm{RCS}, v_d, t_s, x_c, y_c, x_p, y_p]
\end{equation}
As in \cite{Lang.2019}, we then use a set of non-empty BEV grid cells $B$ with a fixed number of points per cell $N_p$ to create a dense tensor of size $(F, B, N_p)$. If the number of non-empty BEV grid cells or the number of points per cell is lower or higher than the fixed number, we apply zero-padding or random sampling, respectively. For each point, we then apply a simplified PointNet \cite{Qi.2017} with a 1$\times$1 convolutional layer followed by BatchNorm and ReLU resulting in a tensor of shape $(C, B, N_p)$, before a max operation over the points per cell reduces the dimension to $(C, B)$. We then map the $C$-dimensional features back to their position on the BEV grid. Finally, the same convolutional backbone as for RadarGridMap is applied.

\subsection{Camera-only baselines}

We selected various camera-only baselines to showcase the plug-in character of our radar fusion module. In this section, we list details for the camera baselines we examined. A compact overview of the modules is given in Table \ref{table:config}.

\begin{table}[bt]
\captionsetup{font=small}
\centering
\caption{Camera-only network configurations. For FPN-LSS, SECOND-FPN and Generalized ResNet, we indicate the number of output channels of the module. For LSS and MatrixVT, we indicate the number of channels and the resolution of the output BEV feature in meters. MF denotes multi-frame temporal fusion.} 
\label{table:config}
\vspace{1mm}
{\scriptsize
\begin{tabular}{lcccccc}
\hline
Model                           & Camera Encoder       & View Transf.                  & BEV Encoder             & Head               & Input Res.         & MF         \\ \hline
\multirow{2}{*}{BEVDet \cite{Huang.2021}}    & Swin-Tiny & LSS & Gen. ResNet-512 & \multirow{2}{*}{CenterPoint} & \multirow{2}{*}{704x256} & \multirow{2}{*}{} \\
                           & FPN-LSS-256          &    80-0.4-0.4                   & FPN-LSS-256             &                              &                          &                    \\ \hline
\multirow{2}{*}{BEVDepth \cite{Li.2022b}}  & ResNet-50            & LSS & ResNet-18               & \multirow{2}{*}{CenterPoint} & \multirow{2}{*}{704x256} & \multirow{2}{*}{\checkmark} \\
                           & SECOND-FPN-128       &  80-0.8-0.8                               & SECOND-FPN-64           &                              &                          &                    \\ \hline
\multirow{2}{*}{BEVStereo \cite{Li.2022c}} & ResNet-50            & LSS & ResNet-18               & \multirow{2}{*}{CenterPoint} & \multirow{2}{*}{704x256} & \multirow{2}{*}{\checkmark} \\
                           & SECOND-FPN-128       &   80-0.8-0.8                              & SECOND-FPN-64           &                              &                          &                    \\ \hline
\multirow{2}{*}{MatrixVT \cite{Zhou.2022c}}  & ResNet-50            & MatrixVT & ResNet-18               & \multirow{2}{*}{CenterPoint} & \multirow{2}{*}{704x256} & \multirow{2}{*}{} \\
                           & SECOND-FPN-128       &   80-0.8-0.8                              & SECOND-FPN-64           &                              &                          &             \\ \hline      
\end{tabular}
}
\vspace{-1mm}
\end{table}

\subsubsection{BEVDet}
BEVDet \cite{Huang.2021} is the first network that uses BEV-based features for object detection on the nuScenes dataset. First, high-level features from each of the $N_i$ input images of shape $(H_i,W_i)$ are extracted separately.
To this end, a SwinTransformer-Tiny \cite{Liu.2021} backbone network outputs multi-scale feature maps, which are then processed using the feature pyramid network from \cite{Philion.2020}, FPN-LSS, which upsamples the low resolution feature maps to match the high resolution map, concatenates them and runs them through a ResNet \cite{He.2016} block. 
This leads to a feature map of shape $(N_i,H_i/8,W_i/8,C)$ with $C$ feature channels. Then, a $1x1$ convolution followed by a Softmax is used to predict a depth classification into $D$ pre-defined depth bins. An outer product across the feature and depth classification channels creates a very large tensor of shape $(N_i,H_i/8,W_i/8,D,C)$. Using the intrinsic and extrinsic calibration matrices of each camera, this tensor can be unprojected into a pseudo-pointcloud. The vertical dimension of this pseudo-pointcloud is then collapsed by summing up the features from all points that fall into the same cell of a pre-defined BEV grid with shape $(H_{bev},W_{bev})$. We follow the implementation of \cite{Liu.2022}, which provides computationally efficient BEV pooling and uses a heavier view transformer to enable more accurate depth estimation, which is important to associate the features from radar and camera.
To further encode the BEV features, the generalized ResNet \cite{He.2016} structure from \cite{Philion.2020} followed again by FPN-LSS is used.

\subsubsection{BEVDepth}
BEVDepth \cite{Li.2022b} uses a similar structure to BEVDet \cite{Huang.2021} but aims to achieve a more accurate depth estimation. To this end, the single convolutional layer for depth estimation in BEVDet \cite{Huang.2021} is replaced with a camera-aware DepthNet module, that concatenates and flattens the camera's intrinsic and extrinsic calibration parameters and uses an MLP to rescale them to match the dimension of the image features $C$. This calibration vector is used to re-weight the image features using a Squeeze-and-Excitation module \cite{Hu.2018}. During training, BEVDepth \cite{Li.2022b} further uses the depth value from projected LiDAR points on the image to directly supervise the depth estimation via binary cross-entropy loss. 
We use the released configuration, which encodes the images with a ResNet-50 \cite{He.2016} backbone followed by the feature pyramid net from \cite{Yan.2018}, SECOND-FPN, which concatenates upsampled multi-scale feature maps.
The BEV encoder uses a ResNet-18 again followed by SECOND-FPN. The view transformer has a lower resolution than the one we use for BEVDet \cite{Huang.2021}. It also uses a multi-frame fusion with one previous keyframe, thus two images from each camera taken 500\,ms apart are used to create the BEV features, resulting in more accurate velocity estimation.

\subsubsection{BEVStereo}
BEVStereo \cite{Li.2022c} builds upon BEVDepth \cite{Li.2022b} and uses the same configuration for most modules. In addition to the monocular depth estimation with DepthNet, it introduces a temporal stereo depth estimation module, which is based on multi-view-stereo \cite{Yao.2018}. To this end, for each pixel in the current image feature map, several depth candidates are predicted and used to retrieve corresponding features from the previous image features using a homography warping. The confidence of each candidate is evaluated based on the current and previous feature's similarity and used to iteratively optimize the depth candidates. After three iterations, the depth candidates are used to construct the stereo depth. Since the stereo depth is not viable for pixels that do not have corresponding pixels in the previous image, a convolutional WeightNet module is used to combine the monocular depth estimations from the current and previous image with the stereo depth estimation to produce the final depth estimates. 

\subsubsection{MatrixVT}
In MatrixVT \cite{Zhou.2022c}, an alternative view transformer is proposed. The view transformation step of LSS \cite{Philion.2020} is first generalized into matrix operations, which leads to a very large and sparse feature transportation tensor of shape $(H_{bev},W_{bev},N_i,H_i/8,W_i/8,D)$, which transforms the image feature tensor with depth estimates to the BEV features. To combat this, the image feature and the dense depth prediction are compressed along the vertical dimension, resulting in prime feature and prime depth matrices. This is feasible due to the low response variance in the height dimension of the image. To further reduce its sparsity, the feature transportation tensor is orthogonally decomposed into a ring tensor, which encodes distance information, and a ray tensor, which encodes directional information. Using efficient mathematical operations, the resulting view transformer achieves lower computational cost and memory requirements. We use a configuration based on BEVDepth \cite{Li.2022b}, which only replaces the view transformer with MatrixVT \cite{Zhou.2022c} and does not use multi-frame fusion.

\subsection{Detection Head and Loss}
All examined camera baselines use the CenterPoint \cite{Yin.2021} detection head, so we can apply the same loss in all architectures. For each class $k$, CenterPoint \cite{Yin.2021} predicts a BEV heatmap with peaks at object center locations. The heatmap is trained using a ground-truth heatmap $y$ filled with Gaussian distributions around ground truth object centers. Given the heatmap score $p_{kij}$ at position $i, j$ in the BEV grid and the ground truth $y_{kij}$, we can compute the Gaussian focal loss \cite{Law.2018} as:
\begin{equation}
    L_{hm} = -\frac{1}{N_o} \sum_{k=1}^{K} \sum_{i=1}^{H_{bev}} \sum_{j=1}^{W_{bev}} 
    \begin{cases}
        (1-p_{kij})^\alpha \log{(p_{kij})} & \, y_{kij} = 1 \\
        (1-y_{kij})^\beta (p_{kij})^\alpha \log{(1-p_{kij})} & \, \text{otherwise}
    \end{cases}
\end{equation}
where $N_o$ is the number of objects per image, $K$ is the number of classes, $H_{bev}$ and $W_{bev}$ are the height and width of the BEV grid, and $\alpha$ and $\beta$ are hyperparameters.
In addition, CenterPoint \cite{Yin.2021} has regression heads that output all parameters needed to decode 3D bounding boxes: a sub-pixel location refinement, a height above ground, the dimensions, the velocity, and the sine and cosine of the yaw rotation angle. The regression heads are trained with L1 loss.

\section{Experimental Results}
\label{section:4}

In this section, we present our experimental results. We first give some more information on the dataset, evaluation metrics and training settings. We then list quantitative results on the nuScenes validation set to ensure a fair comparison between our proposed fusion networks with their camera-only baselines. We also show results for the nuScenes \cite{Caesar.2019} test benchmark and provide an inference example for a qualitative comparison. Ablative studies, detailed class-wise results, rain and night scene evaluation and more qualitative examples are provided in the supplementary material.

\subsection{Data and metrics}
For our experiments, we need a large-scale, real-world dataset with unmasked camera images, series-production automotive radars and 3D bounding box annotations. To the best of our knowledge, the nuScenes dataset \cite{Caesar.2019} is currently the only dataset that fulfils these requirements. It covers data from six cameras at 12\,Hz, five conventional 2+1D radar sensors at 13\,Hz and one LiDAR at 20\,Hz, as well as 1.4 million 3D bounding boxes annotated at 2\,Hz. We follow the official splits into 700 training, 150 validation and 150 test scenes and reduce the 27 annotated classes to the 10 classes evaluated on the benchmark. We also use the official metrics: the mean average precision (mAP), the true positive metrics covering mean errors for translation (mATE), scale (mASE), orientation (mAOE), velocity (mAVE) and nuScenes attribute (mAAE), as well as the condensed nuScenes detection score (NDS).

\subsection{Quantitative Evaluation}
\label{section:quantitative}

\subsubsection{Training details}
For our radar-camera fusion networks, we adopt the configurations from the camera-only baselines listed in Table~\ref{table:config} to allow for a fair comparison. In addition, we design our radar encoder branch so that the BEV features have the same shape and orientation as the camera BEV features. To increase the point cloud density while limiting the use of outdated data, we choose to aggregate five radar sweeps, which corresponds to 300\,ms of past data. 
The aggregated radar point cloud is still sparse when compared with LiDAR data, so that we can reduce the number of non-empty grid cells and points per grid cell of the BEVFeatureNet drastically with respect to the pillar feature encoding in \cite{Lang.2019}.
We empirically find that setting $B=2000$, $N_p=10$, and $C=32$ is sufficient, which allows for little computational overhead. For the Gaussian focal loss, we follow \cite{Law.2018} and set $\alpha=2$ and $\beta=4$. We train for 20 epochs with an AdamW \cite{Loshchilov.2017} optimizer, a base learning rate of 2e-4 and weight decay of 1e-2.

\subsubsection{NuScenes validation results}
We show the results for our radar-camera fusion networks w.r.t. their camera-only baselines on the nuScenes validation split in Table \ref{table:val}. First, we compare the two radar encoders with our model based on BEVDet. In both cases, the proposed fusion offers significant performance increases, with the mAP increasing up to $24\%$ and the NDS $28\%$. The up to $23\%$ reduced translation error shows how the direct depth measurements provided by the radar can lead to more precise location predictions. The most significant improvement of $55\%$ is achieved for the velocity error, which is enabled by the direct velocity measurement of the radar. This effect also helps determining whether an object is currently moving or stopped, thus reducing the attribute error. The two metrics that remain relatively unaffected by the fusion are scale and orientation error. This is as expected since the sparse radar detections do not help to determine an object's size or orientation. Overall, we observe similar results for the RadarGridMap and the BEVFeatureNet encoder. This demonstrates the effectiveness and modularity of the proposed BEV-based feature fusion. In general, we recommend using the BEVFeatureNet because it requires less hand-crafted input, is more scalable, and achieves slightly better results.

In the second part of Table \ref{table:val}, we use the BEVFeatureNet encoder as a plug-in branch in different camera-only baselines. We observe significant performance increase for all examined architectures, again confirming the modularity of the proposed architecture. There are two potential reasons for the difference in relative performance increase between the camera-only baselines. First, BEVDepth and BEVStereo use temporal fusion and therefore achieve better velocity prediction and overall scores, leading to smaller margins for the radar-camera fusion. Second, we use a BEVDet variant with a heavier view transformer especially designed for fusion on the BEV space. This modification may explain the relatively high performance gains. 

Finally, we also measure the inference latency on an Nvidia RTX 2080 Ti GPU to demonstrate that our fusion approach introduces only small computational overhead due to the efficient radar encoder design.

\begin{table*}[bt]
\captionsetup{font=small}
\centering
\caption{Experimental results for our radar-camera fusion used in different architectures on the nuScenes val split. The inference latency $T$ is measured on a Nvidia RTX 2080 Ti.
*We use the implementation of BEVDet-Tiny with a heavier view transformer from \cite{Liu.2022}.
\textsuperscript{\textdagger}We list the results as reported by the authors.}
\label{table:val}
\vspace{1mm}
{\scriptsize
\resizebox{\textwidth}{!}{%
\begin{tabular}{lllcrrrrrrrr}
\hline
& Cam. model        & Radar model & & mAP$\uparrow$ & NDS$\uparrow$ & mATE$\downarrow$ & mASE$\downarrow$ & mAOE$\downarrow$ & mAVE$\downarrow$ & mAAE$\downarrow$ & $T$[ms] \\ \hline
\cite{Huang.2021} & BEVDet* & None &    & 0.350 & 0.411 & 0.660 & 0.275 & 0.532 & 0.918 & 0.260 & 122 \\
\hdashline
\multirow{2}{*}{Ours} & \multirow{2}{*}{BEVDet*}    & \multirow{2}{*}{\scalebox{1.0}[1.0]{RadarGridMap}}  &            & 0.429 & 0.525 & 0.523 & 0.272 & 0.507 & 0.412 & 0.183 & 132 \\
          &               & & $\Delta_r$ & \possg{23\%} & \possg{28\%} & \possg{-21\%} & \posns{-1\%} & \posns{-5\%} & \possg{-55\%} & \possg{-30\%} & \\
\hdashline
\multirow{2}{*}{Ours} & \multirow{2}{*}{BEVDet*}    & \multirow{2}{*}{\scalebox{1.0}[1.0]{BEVFeatureNet}} &            & 0.434 & 0.525 & 0.511 & 0.270 & 0.527 & 0.421 & 0.182 & 139 \\ 
          &               & & $\Delta_r$ & \possg{24\%} & \possg{28\%} & \possg{-23\%} & \posns{-2\%} & \posns{-1\%} & \possg{-54\%} & \possg{-30\%} & \\
\hline
\cite{Li.2022b}\textsuperscript{\textdagger} & BEVDepth & None & & 0.359 & 0.480 & 0.612 & 0.269 & 0.507 & 0.409 & 0.201 & 132 \\ 
\hdashline
\multirow{2}{*}{Ours} & \multirow{2}{*}{BEVDepth}  & \multirow{2}{*}{\scalebox{1.0}[1.0]{BEVFeatureNet}} &            & 0.405 & 0.521 & 0.542 & 0.274 & 0.512 & 0.309 & 0.181 & 146 \\ 
          &               & & $\Delta_r$ & \possg{13\%} & \possg{9\%} & \possg{-11\%} & \negns{2\%} & \negns{1\%} & \possg{-24\%} & \possg{-10\%} & \\
\hline
\cite{Li.2022c}\textsuperscript{\textdagger} & BEVStereo & None & & 0.372 & 0.500 & 0.598 & 0.270 & 0.438 & 0.367 & 0.190 & 308 \\ 
\hdashline
\multirow{2}{*}{Ours} & \multirow{2}{*}{BEVStereo} & \multirow{2}{*}{\scalebox{1.0}[1.0]{BEVFeatureNet}}            &               & 0.423 & 0.545 & 0.504 & 0.268 & 0.453 & 0.270 & 0.174 & 322 \\ 
& & & $\Delta_r$ & \possg{14\%} & \possg{9\%} & \possg{-16\%} & \posns{-1\%} & \negns{3\%} & \possg{-26\%} & \possg{-8\%} & \\
\hline
\cite{Zhou.2022c} & MatrixVT & None & & 0.319 & 0.400 & 0.669 & 0.281 & 0.494 & 0.912 & 0.238 & 54 \\ 
\hdashline
\multirow{2}{*}{Ours} & \multirow{2}{*}{MatrixVT} & \multirow{2}{*}{\scalebox{1.0}[1.0]{BEVFeatureNet}}  & & 0.386 & 0.495 & 0.549 & 0.275 & 0.539 & 0.423 & 0.193 & 64 \\ 
& & & $\Delta_r$ & \possg{21\%} & \possg{24\%} & \possg{-18\%} & \posns{-2\%} & \negsg{9\%} & \possg{-54\%} & \possg{-19\%} & \\
\hline
\end{tabular}}
}
\vspace{-2mm}
\end{table*}

\subsubsection{NuScenes test results}

\begin{table*}[bt]
\captionsetup{font=small}
\centering
\caption{Experimental results for published radar-camera fusion models on the nuScenes test benchmark.}
\label{table:test}
\vspace{1mm}
{\scriptsize
\begin{tabular}{lccccccc}
\hline
Model & mAP$\uparrow$ & NDS$\uparrow$ & mATE$\downarrow$ & mASE$\downarrow$ & mAOE$\downarrow$ & mAVE$\downarrow$ & mAAE$\downarrow$ \\ \hline
CenterFusion \cite{Nabati.2021} & 0.326 & 0.449 & 0.631 & 0.261 & 0.516 & 0.614 & 0.115 \\
CRAFT \cite{Kim.2022} & 0.411 & 0.523 & 0.467 & 0.268 & 0.456 & 0.519 & \textbf{0.114} \\
TransCAR \cite{Pang.2023} & 0.422 & 0.522 & 0.630 & 0.260 & \textbf{0.383} & 0.495 & 0.121 \\
Ours (BEVDet) & \textbf{0.476} & \textbf{0.567} & \textbf{0.444} & \textbf{0.244} & 0.461 & \textbf{0.439} & 0.128 \\ \hline
\end{tabular}
}
\vspace{-2mm}
\end{table*}

\begin{figure*}[bt]
\captionsetup{font=small}
\centering
\begin{subfigure}{0.49\textwidth}
  \adjincludegraphics[trim={.08\width} 0 {.08\width} 0,clip,width=\textwidth]{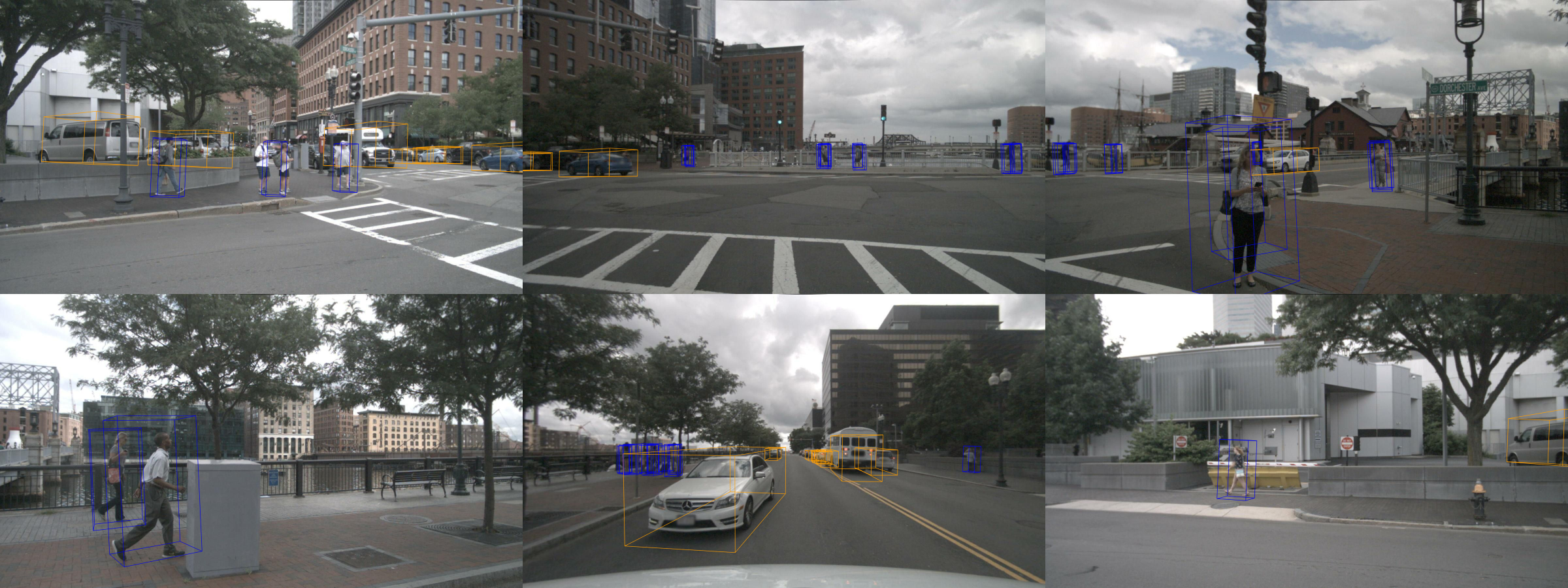}
  \includegraphics[width=\textwidth]{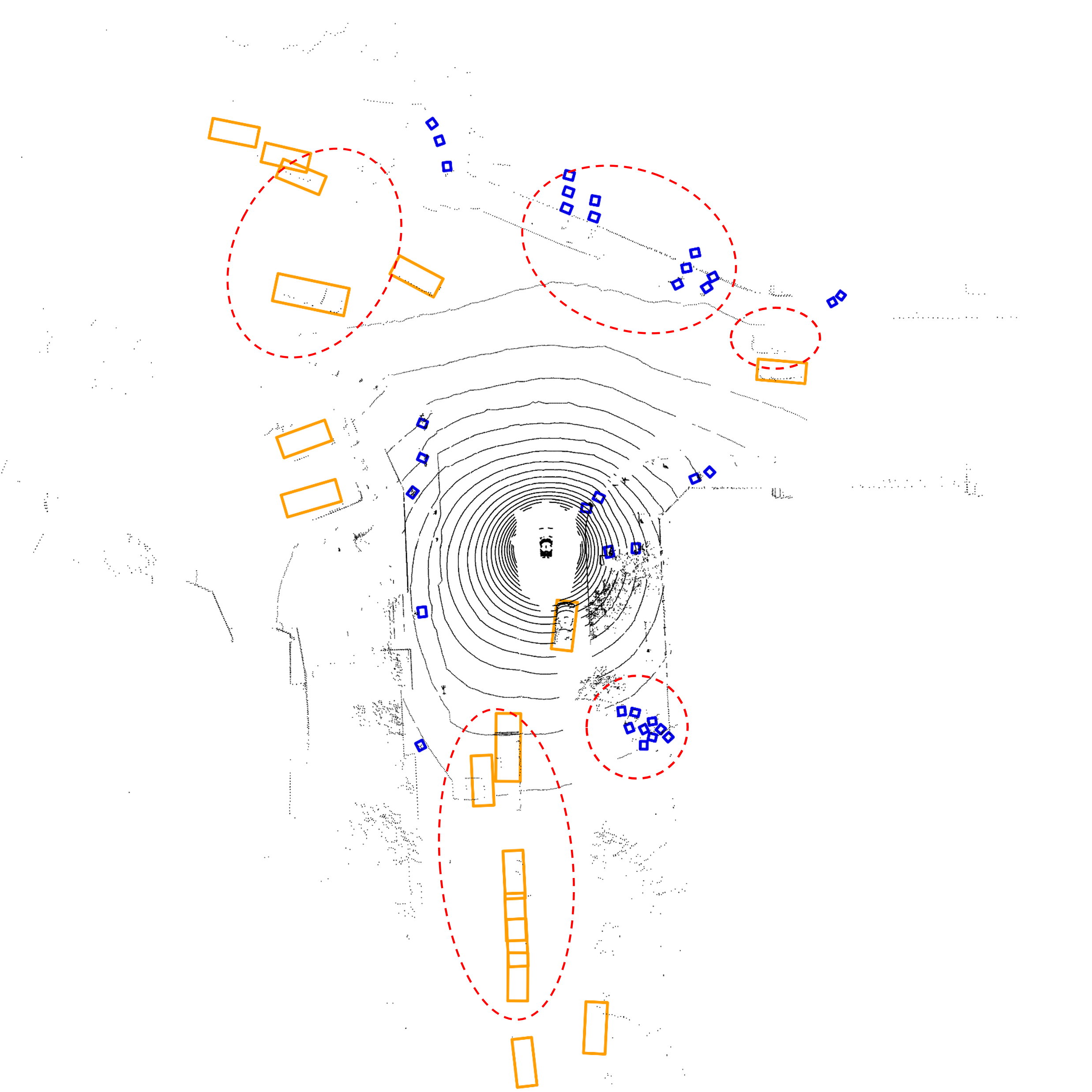}
  \caption{BEVDet}
  \end{subfigure}
  \begin{subfigure}{0.49\textwidth} 
  \adjincludegraphics[trim={.08\width} 0 {.08\width} 0,clip,width=\textwidth]{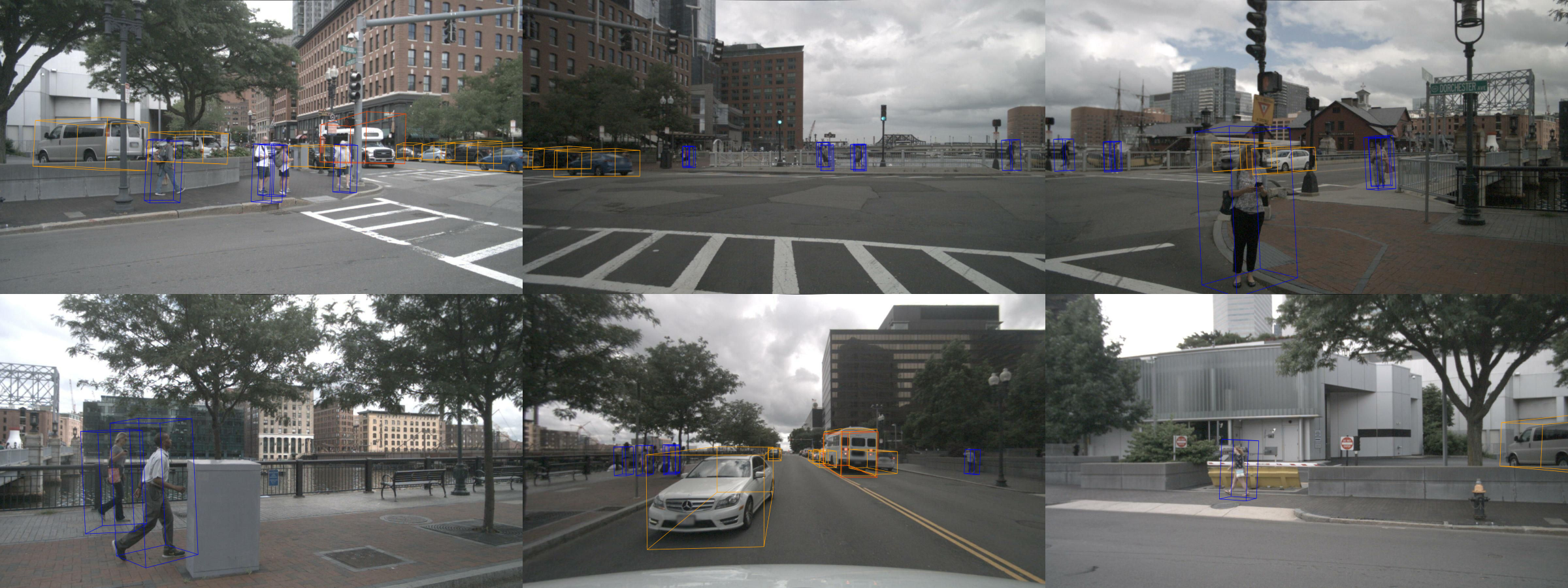}
  \includegraphics[width=\textwidth]{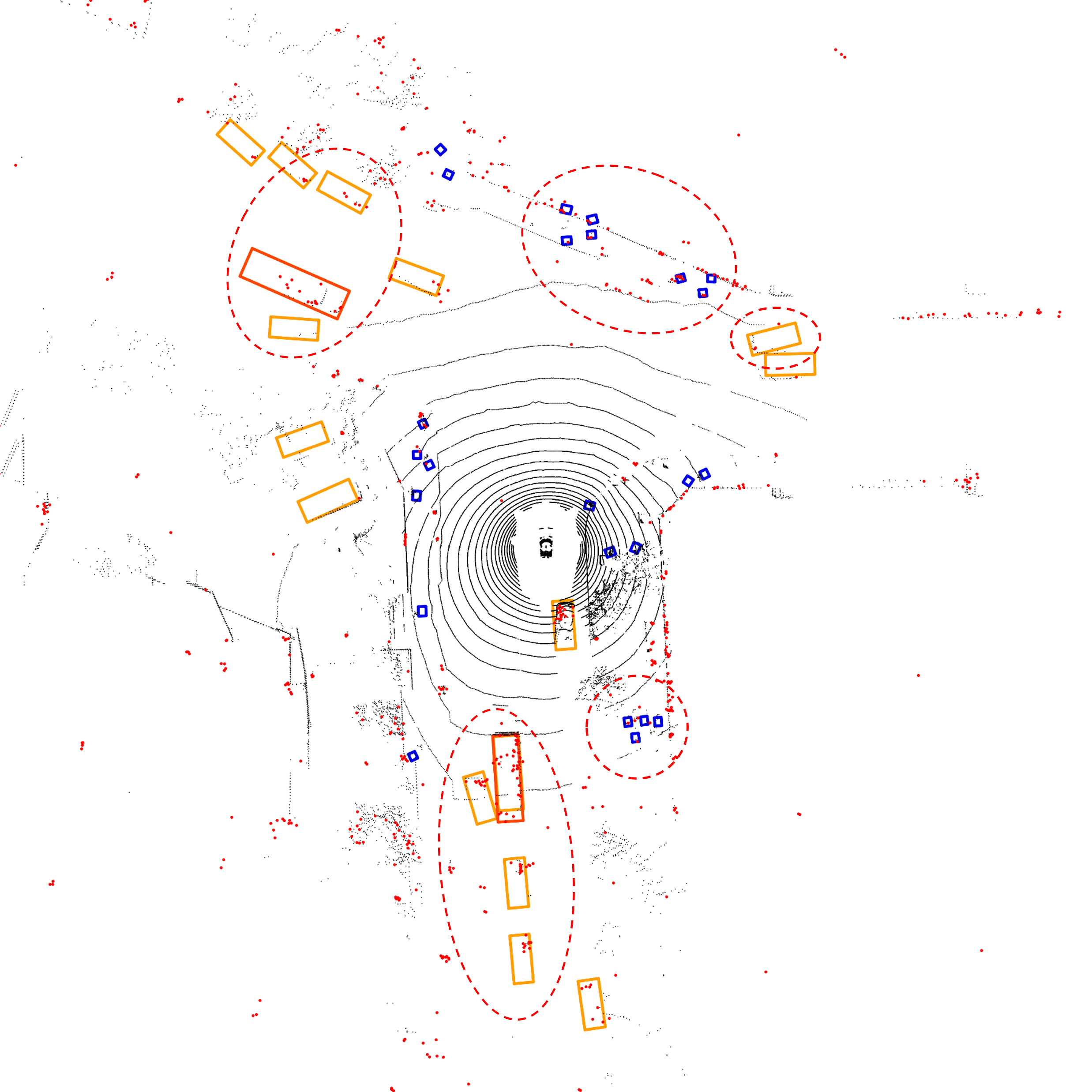}
  \caption{Proposed RC-BEVFusion}
  \end{subfigure}
  \caption{Inference example at daytime. Predicted 3D bounding boxes projected to all six cameras (top) and BEV plane (bottom) with LiDAR points (black) and radar points (red) for reference. Our proposed fusion network more accurately detects pedestrians 
  and vehicles 
  (s. dashed red ellipses).}
  \label{fig:quali_day}
  \vspace{-2mm}
\end{figure*}

For the nuScenes test benchmark, we show our results compared to other published radar-camera fusion models in Table~\ref{table:test}. We note that many methods tune their models for the benchmark submission by enlarging their network and the input image resolution and by using test time augmentation. All of these techniques trade off smaller performance gains for large computational cost and are therefore not helpful in an automotive application in which fast decisions are required. For instance, the authors of BEVDet achieve 15.6 frames per second with the tiny configuration similar to ours, while the base configuration used for the benchmark achieves only 1.9 frames per second \cite{Huang.2021}. We therefore combat this trend and only retrain our model with scenes from the training and validation set for the test submission. As shown in Table~\ref{table:test}, our proposed RC-BEVFusion with BEVFeatureNet and BEVDet significantly outperforms previously published radar-camera fusion networks in all metrics except the orientation error, even without tuning for the benchmark, while achieving 7.2 frames per second. A key advantage of our architecture compared to existing methods \cite{Nabati.2021, Kim.2022, Pang.2023}, is that radar and camera features can equally contribute to the final detections, allowing the model to detect objects that might be missed by each single modality.

\subsection{Qualitative Evaluation}

To get a better impression of the difference in object detection performance, we present an inference example for BEVDet \cite{Huang.2021} and RC-BEVFusion based on BEVFeatureNet and BEVDet in Figure~\ref{fig:quali_day}. We show the camera-only inference results and radar-camera fusion results in the left and right subfigure, respectively. We show the full surround view with all six cameras, the top row shows the front left, center and right camera, while the bottom row shows the back right, center and left camera. On the bottom, we show a BEV perspective with LiDAR points in black for reference and radar points in red for the fusion network only. In each perspective, we show the projected 3D bounding boxes predicted by the networks, where the color indicates the class: pedestrians are blue, cars are yellow and trucks are orange.

In the scene, we see a crowded intersection with lots of cars and pedestrians. At first, the visual impression when looking at the camera images is that most objects are well detected. However, comparing the dashed red ellipses in the BEV perspective on the right, we can see that the radar-camera fusion enables much more accurate detection of the pedestrians in the front and back right area, as well as the cars in the front left, front right and back area.

\section{Conclusion}
\label{section:5}
In this work, we have presented a novel radar-camera fusion architecture on the BEV plane. We propose two radar encoders and show that they can be integrated into several camera-based architectures that use BEV features. In our experiments, the proposed radar-camera fusion outperforms the camera-only baselines by a large margin, demonstrating its effectiveness. Without tuning the model for the test submission to avoid unrealistic computational cost, we outperform previously published radar-camera fusion networks. Our qualitative evaluation shows improved localization accuracy at daytime and higher recall at nighttime.
In future work, we want to study the potential of BEV-based radar-camera fusion with the high-resolution, 3+1D radar sensors appearing in recently introduced datasets.

\section*{Acknowledgment}
This work is partly funded by the German Federal Ministry for Economic Affairs and Climate Action (BMWK) and partly financed by the European Union in the frame of NextGenerationEU within the project ”Solutions and Technologies for Automated Driving in Town” (FKZ 19A22006P) 
and partly funded by the Federal Ministry of Education and Research Germany in the funding program Photonics Research Germany under the project FUMOS (FKZ 13N16302). 
The authors would like to thank the consortia for the successful cooperation.

%
%
%
\bibliographystyle{splncs04}
\bibliography{041-main}

\newpage
\clearpage

\input{supplementary}

\end{document}

%% file: supplementary.tex
\appendix

\setlength{\tabcolsep}{4pt}

\section{Supplementary Material}
In the supplementary material, we present ablative studies regarding augmentation strategies in Section~\ref{section:A1}. Further, we list additional quantitative results including class-wise evaluation results in Section~\ref{section:A2}. To highlight the advantage of a radar-camera fusion, we provide results of a filtered evaluation for rain and night scenes in Section~\ref{section:A3}, in which radar is especially useful. Finally, additional qualitative examples are provided in Section~\ref{section:A4}.

\subsection{Ablations}
\label{section:A1}

In Table \ref{table:augmentation} we show the impact of different augmentation strategies on our RC-BEVFusion with BEVFeatureNet and BEVDet. As in \cite{Huang.2021}, we use a two-stage augmentation. First, 2D augmentation is applied by rotating, resizing and horizontally flipping the images. In the BEV view transformer, the 2D augmentations are reversed so that the original orientation is retained. Then, 3D augmentations are applied to the BEV features, radar points and ground truth boundings boxes. They include scaling, rotating, as well as horizontal and vertical flipping. Our results show that 3D augmentation is very important, while 2D augmentation helps to improve the results only slightly. This is due to the camera encoder branch being shared across the six cameras, leading to more variety in the images than in the BEV plane. Therefore, the BEV encoder is more prone to overfitting and requires more augmentation \cite{Huang.2021}.  

\begin{table}[hbpt]
\captionsetup{font=small}
\centering
\caption{Ablation study for 2D and 3D augmentations. All experiments conducted with the model based on BEVFeatureNet and BEVDet.}
\label{table:augmentation}
{\scriptsize
\begin{tabular}{cccccc}
\hline 
3D & 2D & mAP$\uparrow$ & NDS$\uparrow$ & mATE$\downarrow$ & mAVE$\downarrow$ \\ \hline 
& & 0.380 & 0.453 & 0.595 & 0.676 \\
\checkmark & & 0.419 & 0.513 & 0.520 & 0.478 \\
\checkmark & \checkmark & \textbf{0.434} & \textbf{0.525} & \textbf{0.511} & \textbf{0.421} \\ \hline          
\end{tabular}
}
\end{table}

Further, we examine two potential augmentation strategies for our RadarGridMap encoder, which have been proposed based on a similar setup in \cite{Scheiner.2021}. To combat the sparsity of the radar grid, a blurring filter spreads values from the reference cell to neighboring cells depending on the number of detections per reference cell. Further, the authors find that the distribution of compensated Doppler values is heavy-sided towards zero and introduce a Doppler skewing function to spread the distribution for values close to zero. The results are shown in Table~\ref{table:encoder}. The grid mapping variant without extra augmentation works quite well, confirming the effectiveness of the approach. The blurring filter has little impact on the results and thus is not deemed necessary. The Doppler skewing function leads to higher velocity errors and should therefore be omitted. 

\begin{table}[hbpt]
\captionsetup{font=small}
\centering
\caption{Ablation study for augmentation strategies of the RadarGridMap encoder: a blurring filter (BF) and a Doppler skewing (DS) technique. All experiments conducted with the model based on BEVDet.}
\label{table:encoder}
{\scriptsize
\begin{tabular}{cccccc}
\hline
BF & DS & mAP$\uparrow$ & NDS$\uparrow$ & mATE$\downarrow$ & mAVE$\downarrow$ \\ \hline
 & & \textbf{0.429} & \textbf{0.525} & 0.523 & \textbf{0.412} \\
    \checkmark &  & 0.424 & 0.524 & 0.517 & 0.439 \\
    & \checkmark & 0.425 & 0.508 & 0.523 & 0.482 \\
    \checkmark & \checkmark & 0.423 & 0.487 & \textbf{0.516} & 0.650 \\ \hline   
\end{tabular}
}
\end{table}

\subsection{Class-wise results}
\label{section:A2}

\begin{table}[hbpt]
\captionsetup{font=small}
\centering
\caption{Experimental class-wise results for our radar-camera fusion used in different architectures on the three most common dynamic classes of the nuScenes val split.
*We use the implementation of BEVDet-Tiny with a heavier view transformer from \cite{Liu.2022}.
\textsuperscript{\textdagger}We list the results as reported by the authors.}
\label{table:class_wise}
{\small
\resizebox{\textwidth}{!}{%
\begin{tabular}{llllccccccc}
\hline
 & Cam. model & Radar model & Class & & AP$\uparrow$ & ATE$\downarrow$ & ASE$\downarrow$ & AOE$\downarrow$ & AVE$\downarrow$ & AAE$\downarrow$ \\
\multirow{3}{*}{\cite{Huang.2021}} & \multirow{3}{*}{BEVDet*} & \multirow{3}{*}{None} & car & & 0.538 & 0.529 & 0.157 & 0.128 & 0.992 & 0.232 \\
 & & & ped. & & 0.377 & 0.700 & 0.304 & 1.383 & 0.872 & 0.759 \\
 & & & truck & & 0.290 & 0.679 & 0.209 & 0.165 & 0.911 & 0.252 \\
\hdashline
\multirow{6}{*}{Ours} & \multirow{6}{*}{BEVDet*} & \multirow{6}{*}{BEVFeatureNet} & \multirow{2}{*}{car} & & 0.700 & 0.315 & 0.156 & 0.106 & 0.395 & 0.196 \\
 & & & & $\Delta_r$ & \possg{30\%} & \possg{-40\%} & \posns{-1\%} & \possg{-17\%} & \possg{-60\%} & \possg{-16\%} \\
 & & & \multirow{2}{*}{ped.} & & 0.468 & 0.528 & 0.300 & 1.016 & 0.701 & 0.329 \\
 & & & & $\Delta_r$ & \possg{24\%} & \possg{-25\%} & \posns{-1\%} & \possg{-27\%} & \possg{-20\%} & \possg{-57\%} \\
 & & & \multirow{2}{*}{truck} & & 0.405 & 0.493 & 0.206 & 0.131 & 0.329 & 0.202 \\
 & & & & $\Delta_r$ & \possg{40\%} & \possg{-27\%} & \posns{-1\%} & \possg{-21\%} & \possg{-64\%} & \possg{-20\%} \\
\hline
\multirow{3}{*}{\cite{Li.2022b}\textsuperscript{\textdagger}} & \multirow{3}{*}{BEVDepth} & \multirow{3}{*}{None} & car & & 0.559 & 0.475 & 0.157 & 0.112 & 0.370 & 0.205 \\
 & & & ped. & & 0.363 & 0.690 & 0.297 & 0.831 & 0.491 & 0.244 \\
 & & & truck & & 0.270 & 0.659 & 0.196 & 0.103 & 0.356 & 0.181 \\
\hdashline
\multirow{6}{*}{Ours} & \multirow{6}{*}{BEVDepth} & \multirow{6}{*}{BEVFeatureNet} & \multirow{2}{*}{car} & & 0.661 & 0.356 & 0.162 & 0.134 & 0.289 & 0.193 \\
 & & & & $\Delta_r$ & \possg{18\%} & \possg{-25\%} & \negns{3\%} & \negsg{20\%} & \possg{-22\%} & \possg{-6\%} \\
 & & & \multirow{2}{*}{ped.} & & 0.410 & 0.585 & 0.295 & 0.732 & 0.434 & 0.208 \\
 & & & & $\Delta_r$ & \possg{13\%} & \possg{-15\%} & \posns{-1\%} & \possg{-12\%} & \possg{-12\%} & \possg{-15\%} \\
 & & & \multirow{2}{*}{truck} & & 0.332 & 0.563 & 0.214 & 0.162 & 0.254 & 0.198 \\
 & & & & $\Delta_r$ & \possg{23\%} & \possg{-15\%} & \negsg{9\%} & \negsg{57\%} & \possg{-29\%} & \negsg{9\%} \\
\hline
\multirow{3}{*}{\cite{Li.2022c}\textsuperscript{\textdagger}} & \multirow{3}{*}{BEVStereo} & \multirow{3}{*}{None} & car & & 0.567 & 0.457 & 0.156 & 0.104 & 0.343 & 0.204 \\
 & & & ped. & & 0.402 & 0.653 & 0.297 & 0.803 & 0.479 & 0.249 \\
 & & & truck & & 0.299 & 0.650 & 0.205 & 0.103 & 0.321 & 0.197 \\
\hdashline
\multirow{6}{*}{Ours} & \multirow{6}{*}{BEVStereo} & \multirow{6}{*}{BEVFeatureNet} & \multirow{2}{*}{car} & & 0.687 & 0.324 & 0.159 & 0.106 & 0.250 & 0.192 \\
 & & & & $\Delta_r$ & \possg{21\%} & \possg{-29\%} & \negns{2\%} & \negns{2\%} & \possg{-27\%} & \possg{-6\%} \\
 & & & \multirow{2}{*}{ped.} & & 0.469 & 0.530 & 0.295 & 0.694 & 0.413 & 0.197 \\
 & & & & $\Delta_r$ & \possg{17\%} & \possg{-19\%} & \posns{-1\%} & \possg{-14\%} & \possg{-14\%} & \possg{-21\%} \\
 & & & \multirow{2}{*}{truck} & & 0.364 & 0.516 & 0.208 & 0.106 & 0.214 & 0.184 \\
 & & & & $\Delta_r$ & \possg{22\%} & \possg{-21\%} & \negns{1\%} & \negns{3\%} & \possg{-33\%} & \possg{-7\%} \\
\hline
\multirow{3}{*}{\cite{Zhou.2022c}} & \multirow{3}{*}{MatrixVT} & \multirow{3}{*}{None} & car & & 0.517 & 0.529 & 0.162 & 0.155 & 1.049 & 0.221 \\
 & & & ped. & & 0.309 & 0.746 & 0.300 & 1.204 & 0.813 & 0.465 \\
 & & & truck & & 0.244 & 0.713 & 0.213 & 0.154 & 0.917 & 0.219 \\
\hdashline
\multirow{6}{*}{Ours} & \multirow{6}{*}{MatrixVT} & \multirow{6}{*}{BEVFeatureNet} & \multirow{2}{*}{car} & & 0.658 & 0.346 & 0.162 & 0.141 & 0.400 & 0.190 \\
 & & & & $\Delta_r$ & \possg{27\%} & \possg{-35\%} & \posns{0\%} & \possg{-9\%} & \possg{-62\%} & \possg{-14\%} \\
 & & & \multirow{2}{*}{ped.} & & 0.386 & 0.618 & 0.298 & 1.071 & 0.695 & 0.335 \\
 & & & & $\Delta_r$ & \possg{25\%} & \possg{-17\%} & \posns{-1\%} & \possg{-11\%} & \possg{-15\%} & \possg{-28\%} \\
 & & & \multirow{2}{*}{truck} & & 0.320 & 0.547 & 0.214 & 0.133 & 0.337 & 0.201 \\
 & & & & $\Delta_r$ & \possg{31\%} & \possg{-23\%} & \posns{0\%} & \possg{-14\%} & \possg{-63\%} & \possg{-8\%} \\
\hline
\end{tabular}}
}
\end{table}

In addition to the quantitative evaluation in Section~\ref{section:quantitative} and Table~\ref{table:val}, we present some more detailed, class-wise results in Table~\ref{table:class_wise}. To reduce the amount of data, we only list values of the three most common dynamic classes: car, pedestrian, and truck. The results show that the average precision for important classes increased across the board, with an even greater increase for the larger classes that are more easily detectable by radar: car and especially truck. Translation errors decreased most notably for cars, while scale errors remained relatively unchanged, except for a slight decrease for our model based on BEVDepth \cite{Li.2022b}, which may be due to statistical effects. Orientation errors improved for models without temporal fusion, but increased for our model based on BEVDepth, possibly indicating difficulty in estimating the orientation of some additionally detected cars and trucks. Velocity errors saw a significant reduction, especially for cars and trucks and for models without temporal fusion. Even with temporal fusion, there was a considerable decrease in velocity error. Finally, the attribute error decreased most for pedestrians, with radar making it easier to determine if a pedestrian is moving or standing.

\subsection{Results for rain and night scenes}
\label{section:A3}

In this section, we want to evaluate the performance of our radar-camera fusion in adverse conditions for the camera. We therefore filter the scene descriptions in the nuScenes \cite{Caesar.2019} validation set for the terms ``rain'' and ``night'', respectively, to obtain 27 rain and 15 night scenes, on which we run the evaluation for BEVDet \cite{Huang.2021} and our corresponding RC-BEVFusion algorithm with BEVFeatureNet. Since not all classes are represented in the rain and night scenes, the averaged metrics across all classes are less meaningful. We therefore again present class-wise results of the three most common dynamic classes: car, pedestrian, and truck. The results in Table~\ref{table:rain_night} show that compared to the overall AP increase across all scenes listed in Table~\ref{table:class_wise}, there were higher improvements for rain and especially for night scenes. We conclude that the camera-only model struggles in these adverse conditions, particularly in detecting pedestrians and trucks at night. This is where the proposed radar-camera fusion can add the most value. Note that the true positive metrics for pedestrians and trucks at night should be treated with caution due to the low number of matches for the camera-only model. As discussed above, we again observe significant decreases in translation, velocity, and attribute errors, while the scale error remains relatively unchanged. This time, there is also a considerable improvement in orientation error. This finding suggests that the camera-only model struggles to accurately predict orientation in these adverse conditions, and further emphasizes the potential of radar-camera fusion in such scenarios.

\begin{table}[hpbt]
\captionsetup{font=small}
\centering
\caption{Experimental class-wise results for our radar-camera fusion based on BEVDet on the three most common dynamic classes of the nuScenes val split, filtered by rain and night scenes, respectively.
*We use the implementation of BEVDet-Tiny with a heavier view transformer from \cite{Liu.2022}.}
\label{table:rain_night}
{\small
\resizebox{\textwidth}{!}{%
\begin{tabular}{lllllccccccc}
\hline
Split & & Cam. model & Radar model & Class & & AP$\uparrow$ & ATE$\downarrow$ & ASE$\downarrow$ & AOE$\downarrow$ & AVE$\downarrow$ & AAE$\downarrow$ \\
\multirow{9}{*}{Rain} & \multirow{3}{*}{\cite{Huang.2021}} & \multirow{3}{*}{BEVDet*} & \multirow{3}{*}{None} & car & & 0.517 & 0.548 & 0.158 & 0.133 & 0.693 & 0.151 \\
& & & & ped. & & 0.218 & 0.748 & 0.360 & 1.679 & 1.043 & 0.725 \\
& & & & truck & & 0.276 & 0.751 & 0.216 & 0.153 & 0.479 & 0.120 \\
\cdashline{2-12}
& \multirow{6}{*}{Ours} & \multirow{6}{*}{BEVDet*} & \multirow{6}{*}{BEVFeatureNet} & \multirow{2}{*}{car} & & 0.723 & 0.304 & 0.160 & 0.121 & 0.277 & 0.141 \\
& & & & & $\Delta_r$ & \possg{40\%} & \possg{-45\%} & \negns{1\%} & \possg{-9\%} & \possg{-60\%} & \possg{-7\%} \\
& & & & \multirow{2}{*}{ped.} & & 0.338 & 0.516 & 0.360 & 1.011 & 0.849 & 0.332 \\
& & & & & $\Delta_r$ & \possg{55\%} & \possg{-31\%} & \posns{0\%} & \possg{-40\%} & \possg{-19\%} & \possg{-54\%} \\
& & & & \multirow{2}{*}{truck} & & 0.449 & 0.539 & 0.200 & 0.120 & 0.235 & 0.108 \\
& & & & & $\Delta_r$ & \possg{63\%} & \possg{-28\%} & \possg{-7\%} & \possg{-22\%} & \possg{-51\%} & \possg{-10\%} \\
\hline
\multirow{9}{*}{Night} & \multirow{3}{*}{\cite{Huang.2021}} & \multirow{3}{*}{BEVDet*} & \multirow{3}{*}{None} & car & & 0.403 & 0.527 & 0.137 & 0.111 & 1.619 & 0.485 \\
& & & & ped. & & 0.045 & 0.664 & 0.296 & 1.509 & 0.675 & 0.469 \\
& & & & truck & & 0.057 & 0.630 & 0.221 & 0.151 & 2.795 & 0.582 \\
\cdashline{2-12}
& \multirow{6}{*}{Ours} & \multirow{6}{*}{BEVDet*} & \multirow{6}{*}{BEVFeatureNet} & \multirow{2}{*}{car} & & 0.611 & 0.310 & 0.137 & 0.092 & 0.538 & 0.469 \\
& & & & & $\Delta_r$ & \possg{52\%} & \possg{-41\%} & \posns{0\%} & \possg{-17\%} & \possg{-67\%} & \posns{-3\%} \\
& & & & \multirow{2}{*}{ped.} & & 0.191 & 0.262 & 0.283 & 0.810 & 0.592 & 0.037 \\
& & & & & $\Delta_r$ & \possg{324\%} & \possg{-61\%} & \posns{-4\%} & \possg{-46\%} & \possg{-12\%} & \possg{-92\%} \\
& & & & \multirow{2}{*}{truck} & & 0.265 & 0.304 & 0.181 & 0.127 & 0.616 & 0.697 \\
& & & & & $\Delta_r$ & \possg{365\%} & \possg{-52\%} & \possg{-18\%} & \possg{-16\%} & \possg{-78\%} & \possg{20\%} \\
\hline
\end{tabular}}
}
\end{table}

\subsection{Additional qualitative evaluation}
\label{section:A4}
\begin{figure}[hbpt]
\captionsetup{font=small}
\centering
\begin{subfigure}{0.495\textwidth}
  \adjincludegraphics[trim={.08\width} 0 {.08\width} 0,clip,width=\textwidth]{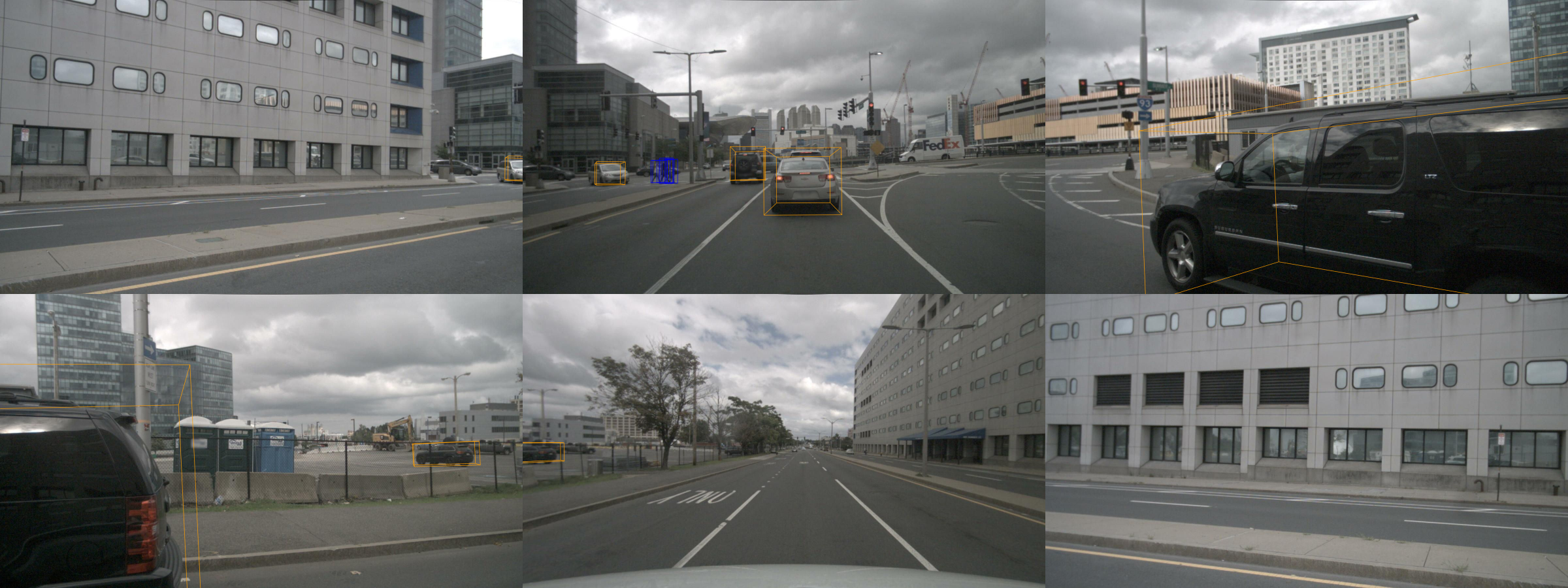}
  \includegraphics[width=0.85\textwidth]{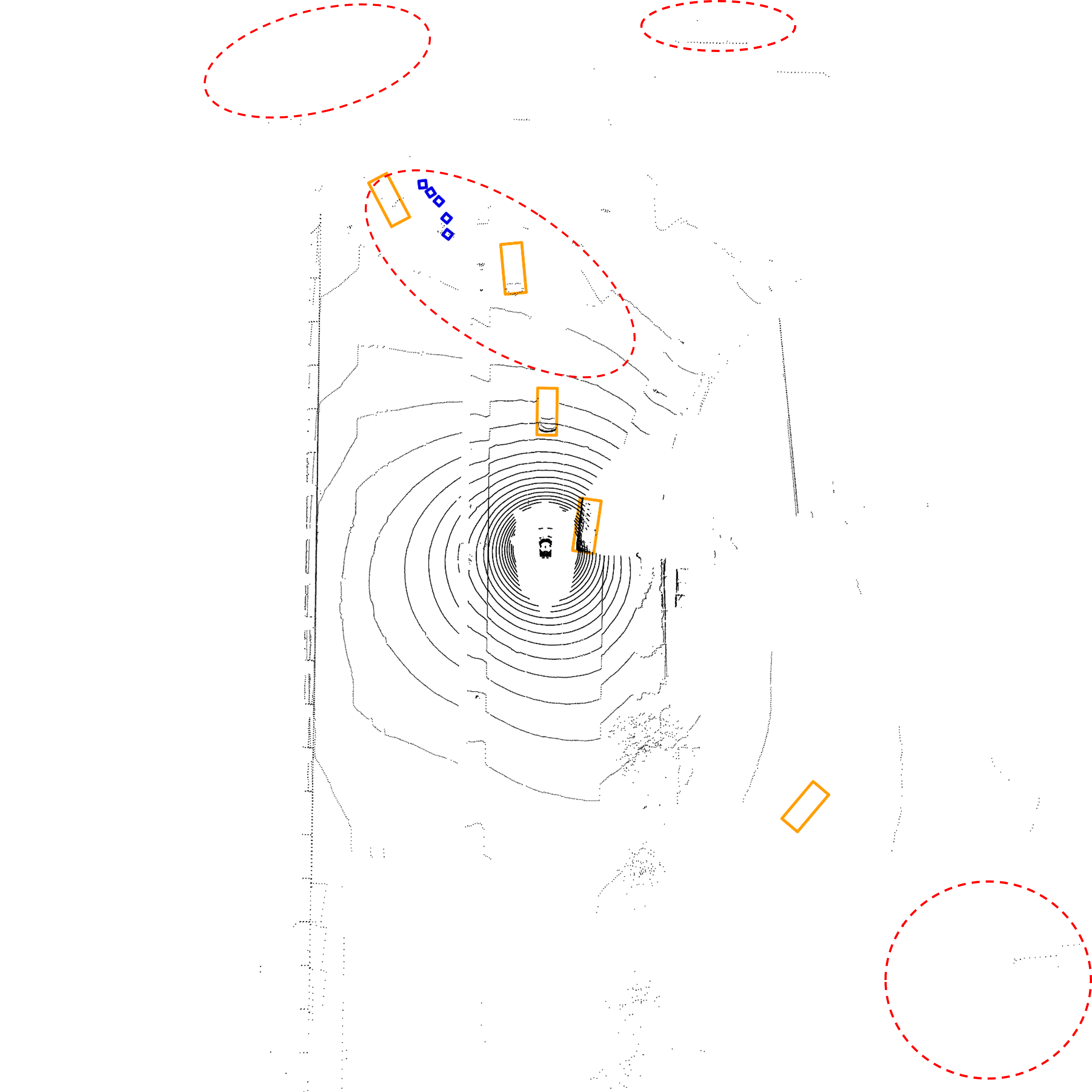}
  \caption{BEVDet}
  \end{subfigure}
  \begin{subfigure}{0.495\textwidth} 
  \adjincludegraphics[trim={.08\width} 0 {.08\width} 0,clip,width=\textwidth]{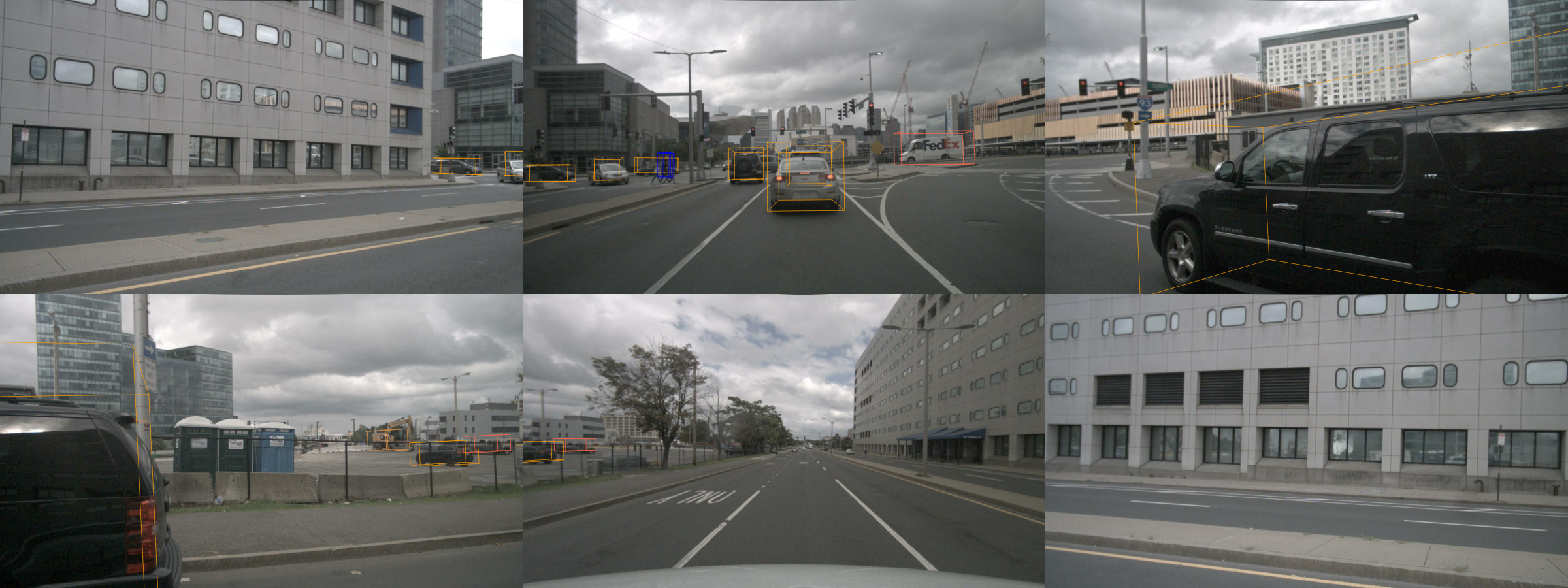}
  \includegraphics[width=0.85\textwidth]{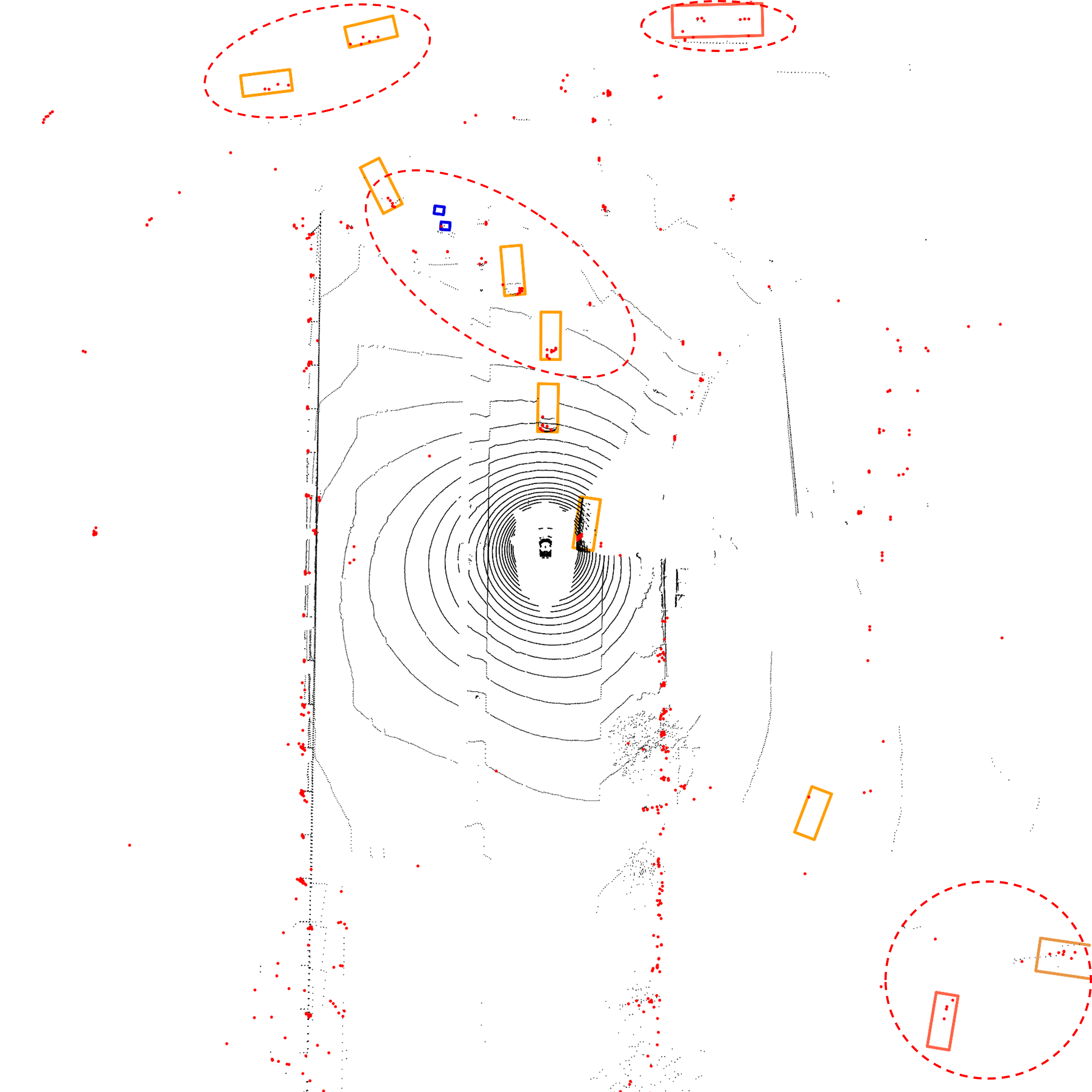}
  \caption{Proposed RC-BEVFusion}
  \end{subfigure}
  \caption{Inference example at daytime. Our network more accurately detects distant cars in the front and back right area as well as an occluded car directly in front.}
  \label{fig:quali_day2}
  \vspace{-2mm}
\end{figure}

\begin{figure}[hbpt]
\captionsetup{font=small}
\centering
\begin{subfigure}{0.495\textwidth}
  \adjincludegraphics[trim={.08\width} 0 {.08\width} 0,clip,width=\textwidth]{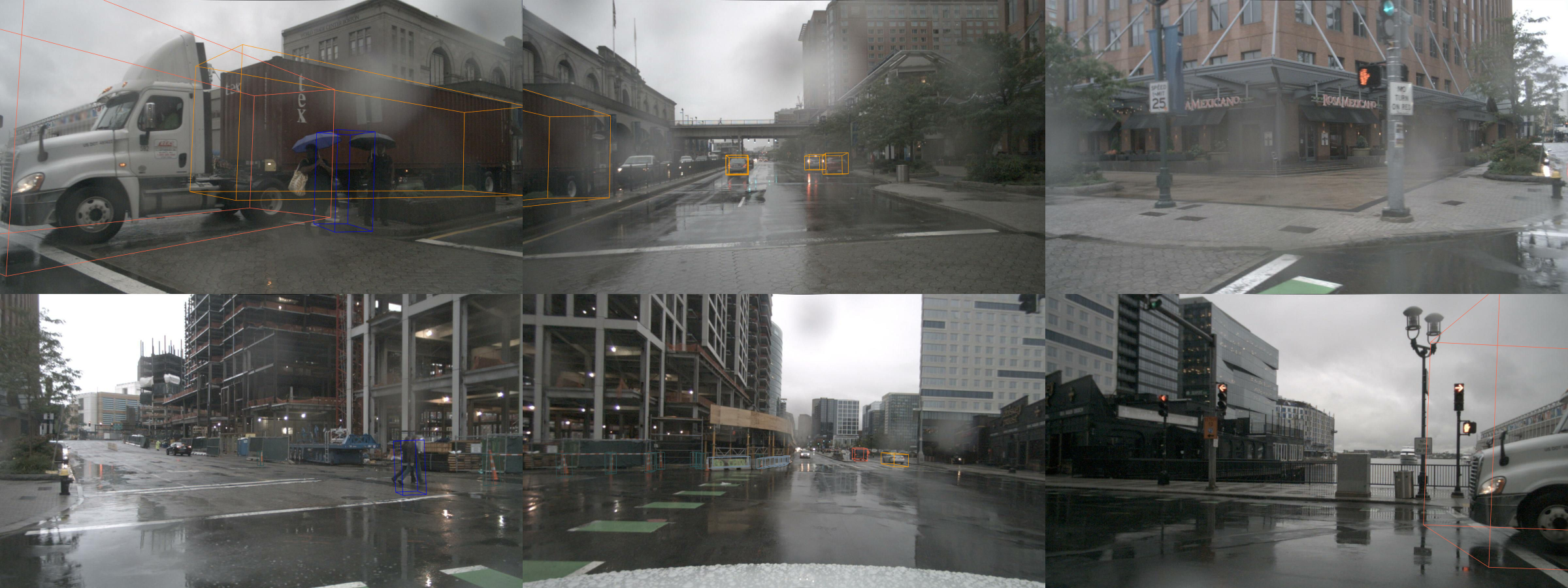}
  \includegraphics[width=0.85\textwidth]{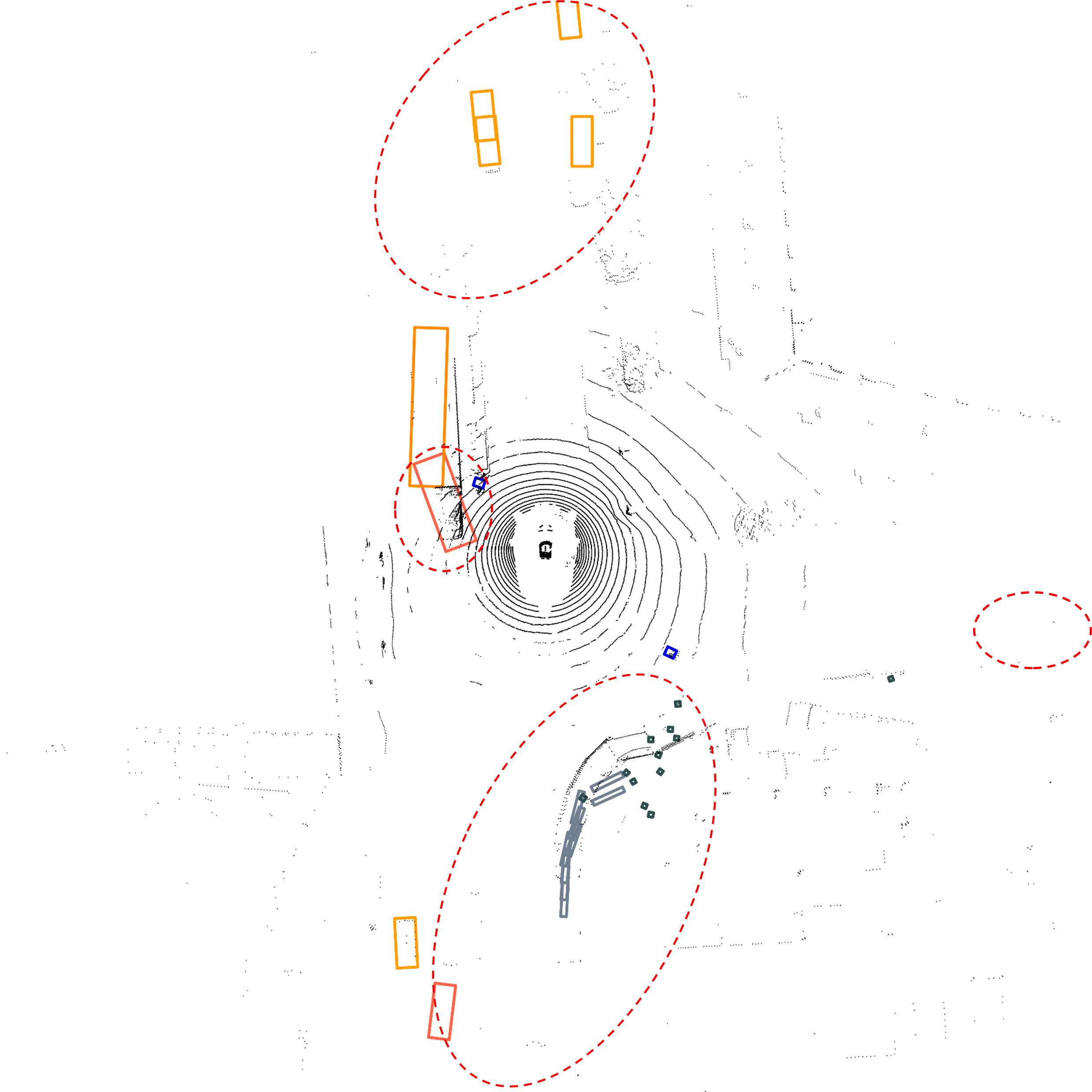}
  \caption{BEVDet}
  \end{subfigure}
  \begin{subfigure}{0.495\textwidth} 
  \adjincludegraphics[trim={.08\width} 0 {.08\width} 0,clip,width=\textwidth]{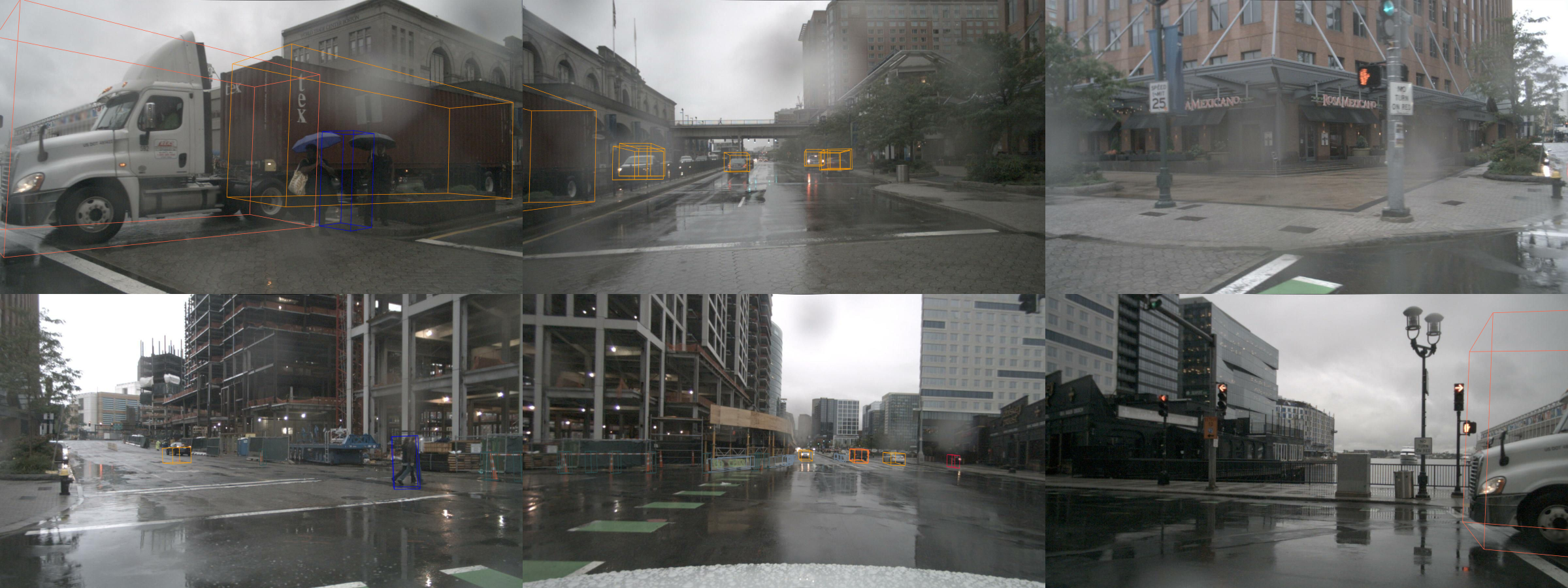}
  \includegraphics[width=0.85\textwidth]{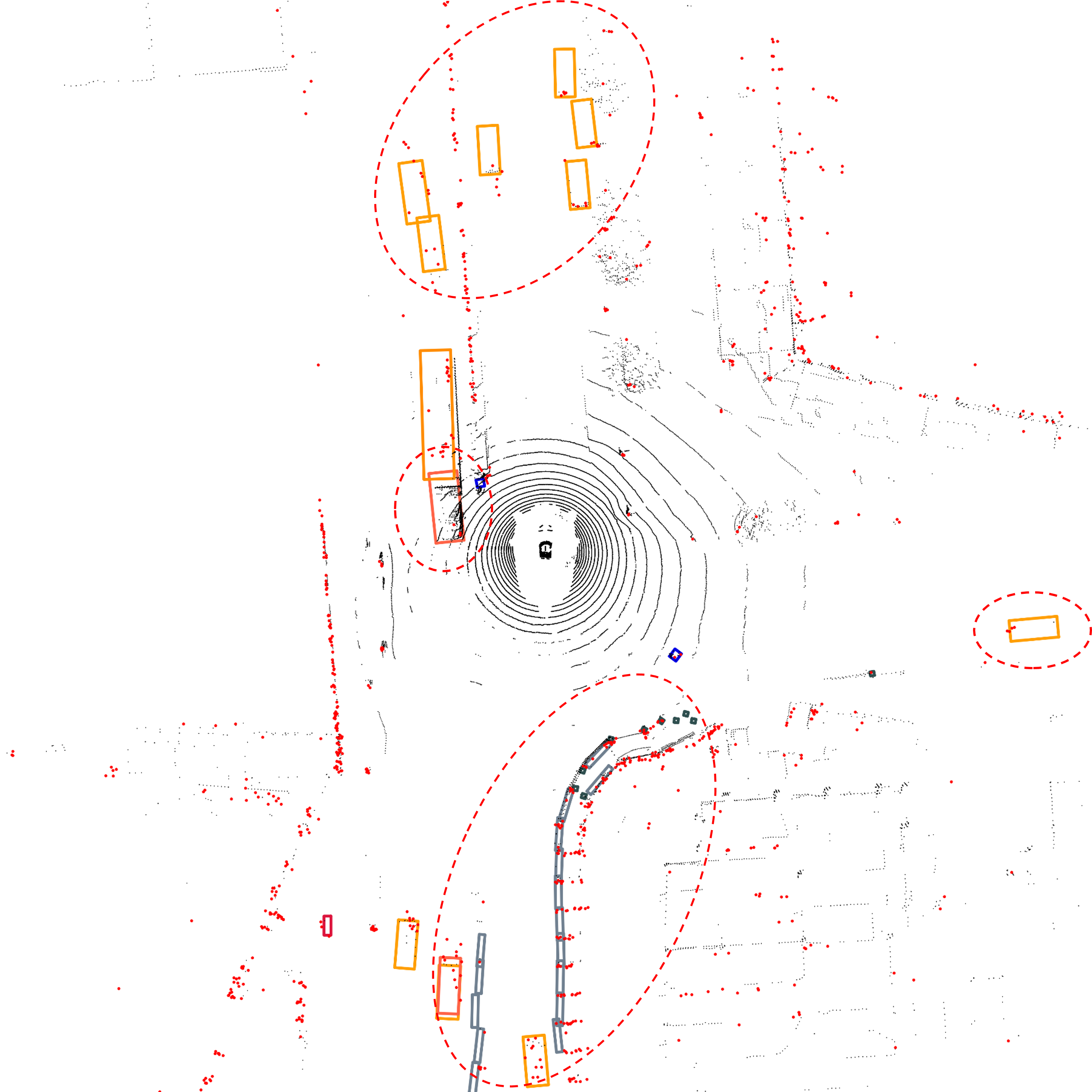}
  \caption{Proposed RC-BEVFusion}
  \end{subfigure}
  \caption{Inference example during rain. Our network has much better overall scene understanding with the road barriers in the back and the cars in front and on the right.}
  \label{fig:quali_rain}
  \vspace{-2mm}
\end{figure}

\begin{figure*}[bt]
\captionsetup{font=small}
\centering
\begin{subfigure}{0.495\textwidth}
  \adjincludegraphics[trim={.08\width} 0 {.08\width} 0,clip,width=\textwidth]{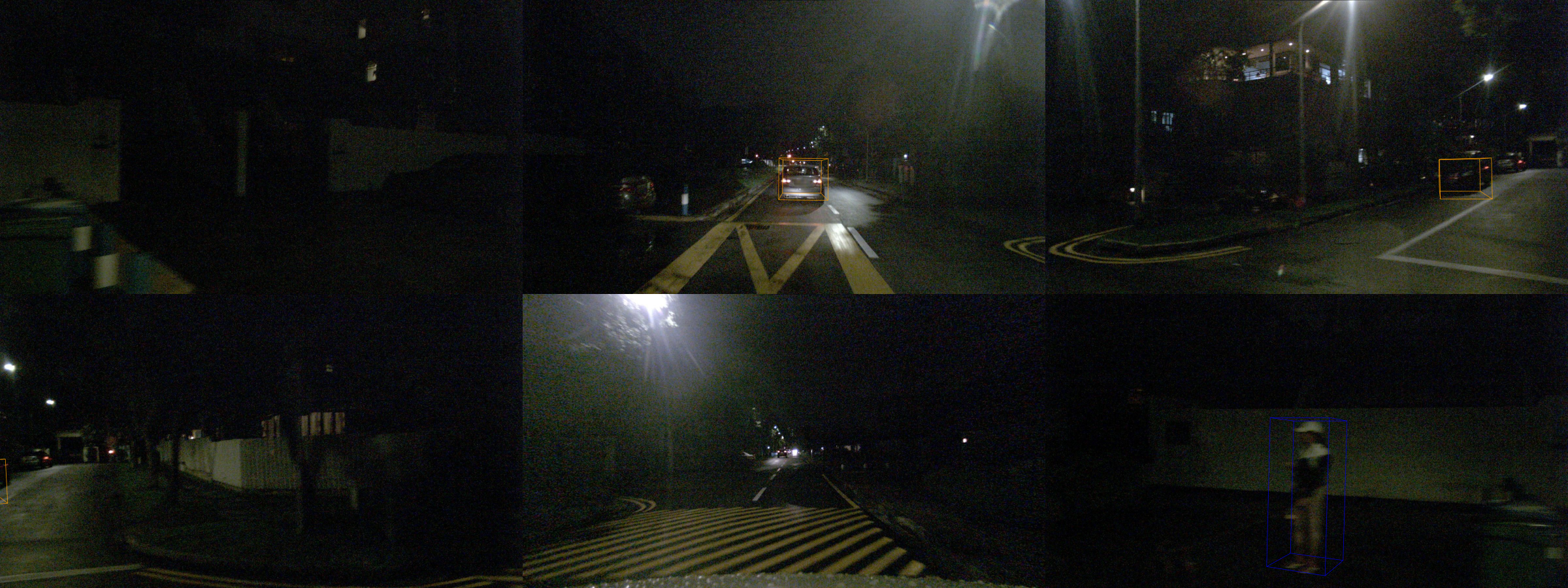}
  \includegraphics[width=\textwidth]{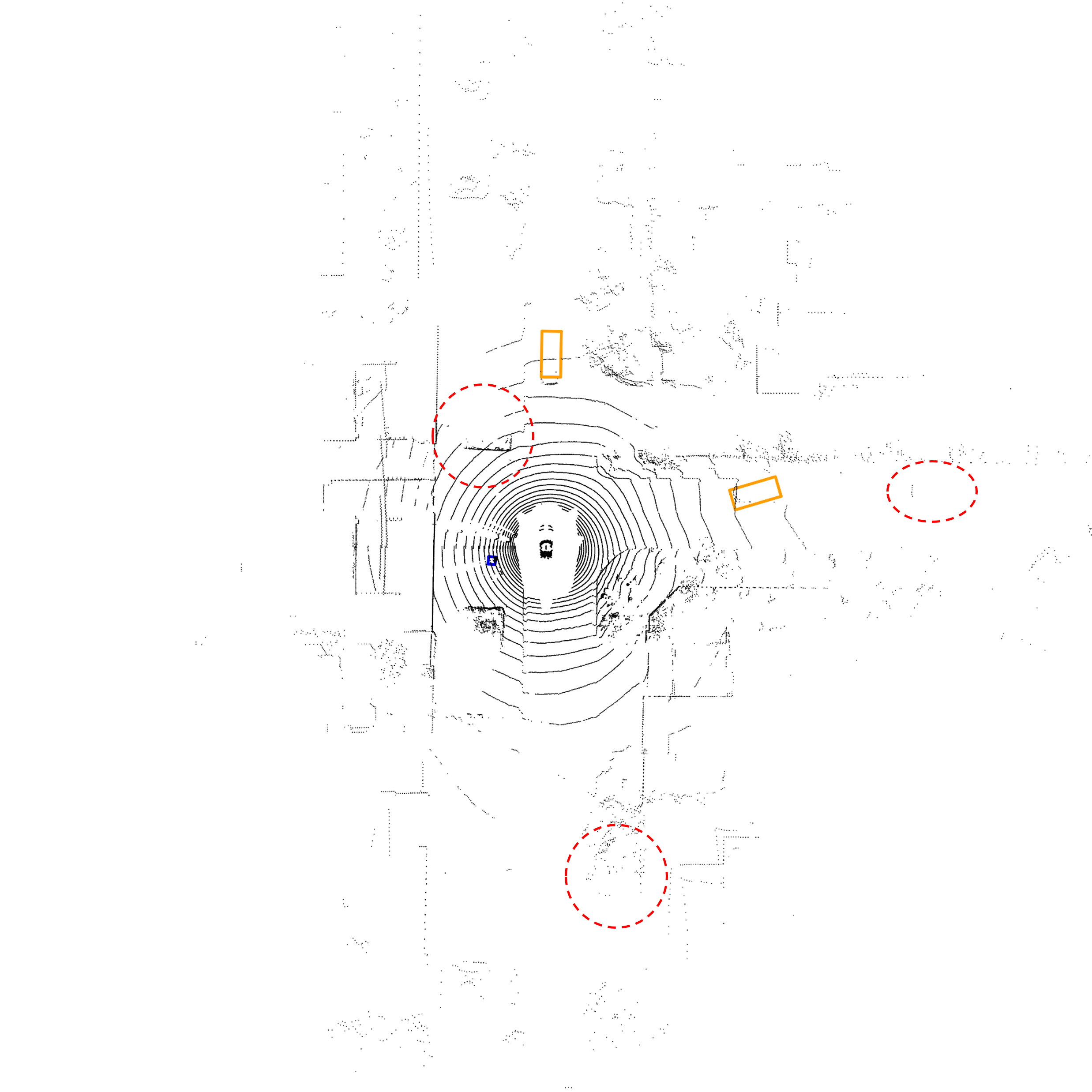}
  \caption{BEVDet}
  \end{subfigure}
  \begin{subfigure}{0.495\textwidth} 
  \adjincludegraphics[trim={.08\width} 0 {.08\width} 0,clip,width=\textwidth]{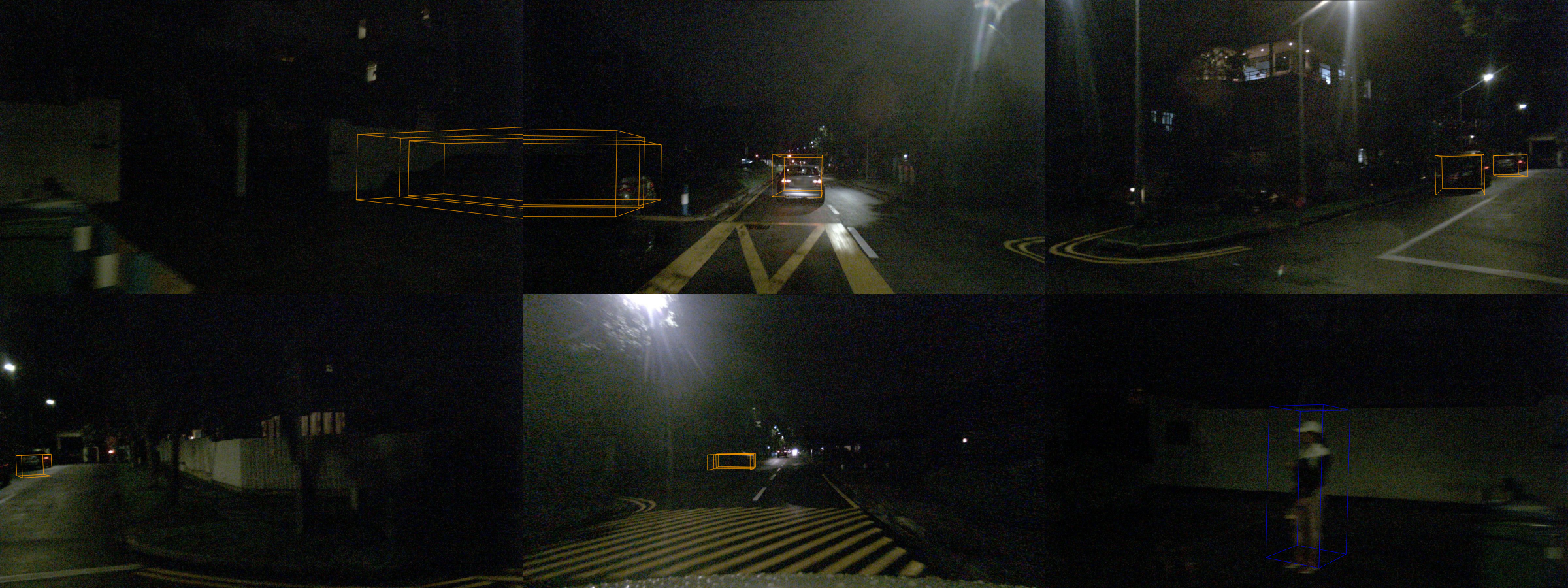}
  \includegraphics[width=\textwidth]{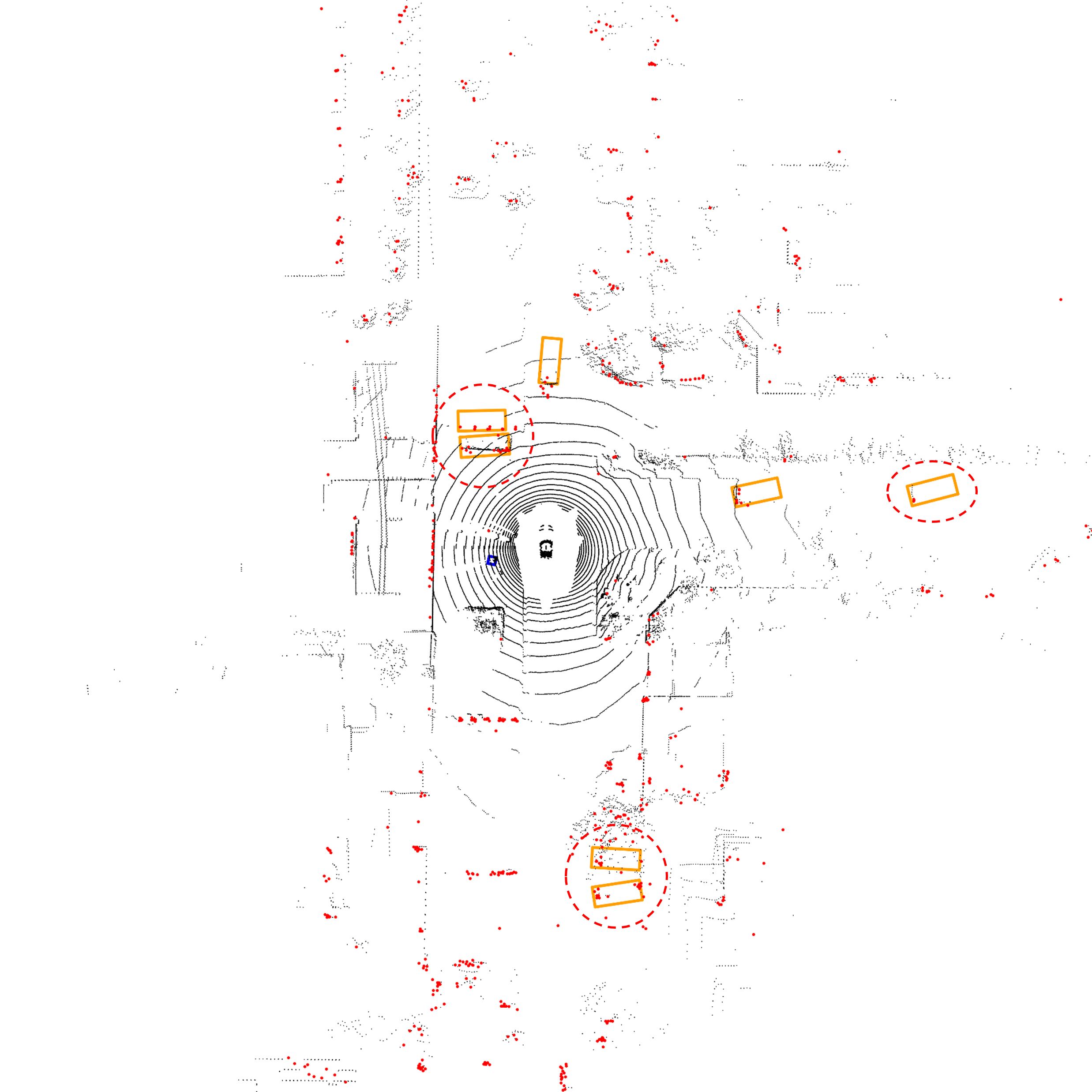}
  \caption{Proposed RC-BEVFusion}
  \end{subfigure}
  \caption{Inference example at nighttime. Our network detects several cars that were missed by the camera-only network due to difficult lighting.}
  \label{fig:quali_night}
  \vspace{-2mm}
\end{figure*}

We provide additional selected qualitative examples for the camera-only baseline BEVDet \cite{Huang.2021} in comparison with our proposed radar-camera fusion model with BEVFeatureNet and BEVDet. Figure~\ref{fig:quali_day2} shows another example at daytime. Our fusion network achieves better performance for long ranges as can be seen with the distant cars in the front and back right area. It also detects an occluded car two vehicles ahead of the ego-vehicle. Figure~\ref{fig:quali_rain} shows an example during rain. As indicated by the quantitative evaluation in the previous section, the network shows a much better overall scene understanding with the barriers in the back area as well as the vehicles in the front area. In addition, the orientation estimation for the truck on the left is improved and an additional vehicle is detected on the right. However, this frame also shows some failure cases that still exist. The two pedestrians on the left are only detected as one by both networks, which are possibly confused by the umbrellas. Also, the fusion network detects an additional parked car in the front right due to matching radar detections, which is a false positive. Figure~\ref{fig:quali_night} shows another example at night with several cars and one pedestrian. We can observe that the camera fails to detect several vehicles, which are difficult to spot given the low light conditions. With our radar-camera fusion, we are able to accurately detect the objects even in difficult lighting conditions.

